\documentclass[11pt]{amsart}
\usepackage{geometry}
\usepackage{graphicx}
\usepackage{amssymb, array, latexsym}

\newcommand{\uu}{{\bf u}}
\newcommand{\vv}{{\bf v}}
\newcommand{\rr}{{\bf r}}
\newcommand{\Estu}{{\mathbb{S}^2}}
\newcommand{\Artri}{{\mathbb{R}^3}}
\newcommand{\LMAX}{{\lambda_\parallel}}
\newcommand{\LMIN}{{\lambda_\perp}}

\title[Spatially regularized reconstruction of fibre orientation distributions ...]{Spatially regularized reconstruction of fibre orientation distributions in the presence of isotropic diffusion}
\author{Q. Zhou, O. Michailovich, and Y. Rathi}

\begin{document}
\maketitle

\begin{abstract}
The connectivity and structural integrity of the white matter of the brain is nowadays known to be implicated into a wide range of brain-related disorders. However, it was not before the advent of diffusion Magnetic Resonance Imaging (dMRI) that researches have been able to examine the properties of white matter {\it in vivo}. Presently, among a range of various methods of dMRI, high angular resolution diffusion imaging (HARDI) is known to excel in its ability to provide reliable information about the local orientations of neural fasciculi (aka fibre tracts). Moreover, as opposed to the more traditional diffusion tensor imaging (DTI), HARDI is capable of distinguishing the orientations of multiple fibres passing through a given spatial voxel. Unfortunately, the ability of HARDI to discriminate between neural fibres that cross each other at acute angles is always limited, which is the main reason behind the development of numerous post-processing tools, aiming at the improvement of the directional resolution of HARDI. Among such tools is spherical deconvolution (SD). Due to its ill-posed nature, however, SD standardly relies on a number of {\it a priori} assumptions which are to render its results unique and stable. In this paper, we propose a different approach to the problem of SD in HARDI, which accounts for the spatial continuity of neural fibres as well as the presence of isotropic diffusion. Subsequently, we demonstrate how the proposed solution can be used to successfully overcome the effect of partial voluming, while preserving the spatial coherency of cerebral diffusion at moderate-to-severe noise levels. In a series of both {\it in silico} and {\it in vivo} experiments, the performance of the proposed method is compared with that of several available alternatives, with the comparative results clearly supporting the viability and usefulness of our approach. 
\end{abstract}


\section{Introduction} 
The connectivity and structural integrity of white matter is nowadays known to be indicative of a wide range of brain-related pathologies. While ``invisible" to alternative means of imaging-based diagnosis, the above information can be elicited from the measurements acquired by means of {\em diffusion Magnetic Resonance Imaging} (dMRI). This fact has triggered an active development of various dMRI methodologies, which has made dMRI into a well-established technique of modern medical imaging \cite{Johansen-Berg:2009sh}. 

At the present time, dMRI encompasses a number of various methodologies and protocols, the most widely acknowledged of which is Diffusion Tensor Imaging (DTI) \cite{Basser:1994pd, Basser:1994ul, Bihan:1986lp, Bihan:2001dp, Mori:2007rw}. It is known, however, that the modelling capacity of DTI is limited due to its reliance on assuming the ensemble averaged diffusion propagator (EAP) to be a unimodal Gaussian. In fact, the above assumption undermines the ability of DTI to provide accurate estimation of the apparent diffusivity of white matter at the locations of crossing, diverging, and kissing neural fibre tracts \cite{Frank:2001mw, Frank:2002bd, Tuch:2002nx}. Alternatively, a parameter-free approach to estimate EAP is provided  by Diffusion Spectral Imaging (DSI) \cite{Wedeen:2005kx}, which allows reconstruction of complex diffusivity profiles under rather general conditions. Unfortunately, practical implementation of DSI entails acquisition of diffusion measurements over a dense Cartesian grid in the $q$-space, which renders the acquisition requirements of DSI beyond the limits of practically admissible. This problem, however, can be alleviated by restricting the diffusion measurements to a relatively small number of concentric shells in the $q$-space. This sampling strategy -- known as Multi-Shell Diffusion Imaging (MSDI) \cite{Descoteaux:2011rt, Menzel:2011fk, Merlet:2010kx} -- has successfully served as a basis for many advanced dMRI methodologies \cite{Descoteaux:2011rt, Wu:2007uq, Jensen:2005ys, Assaf:2004kx, Zhang:2012aa, Assemlal:2008mz, Merlet:2011rt, Ozarslan:2013aa}.

Although the availability of the EAP is generally preferred, in some cases it is sufficient to know the result of its marginalization over the range variable. The resulting probability density is known as the orientation distribution function (ODF), and it quantifies the probability with which water molecules undergo displacement along various spatial directions \cite{Tuch:2003oq,Tuch:2004cr}. On the practical side, a useful approximation of the ODF can be obtained by means of $q$-ball imaging (QBI) \cite{Tuch:2004cr}, which can in turn be based on a single-shell data acquisition scheme, known as High Angular Resolution Diffusion Imaging (HARDI) \cite{Alexander:2005fk, Hess:2006lo, Descoteaux:2007jw, Descoteaux:2006qf, Frank:2002bd, Aganj:2010fk, Barnett:2009ee}.

\begin{figure}[!t]
\centering
\includegraphics[width=4in]{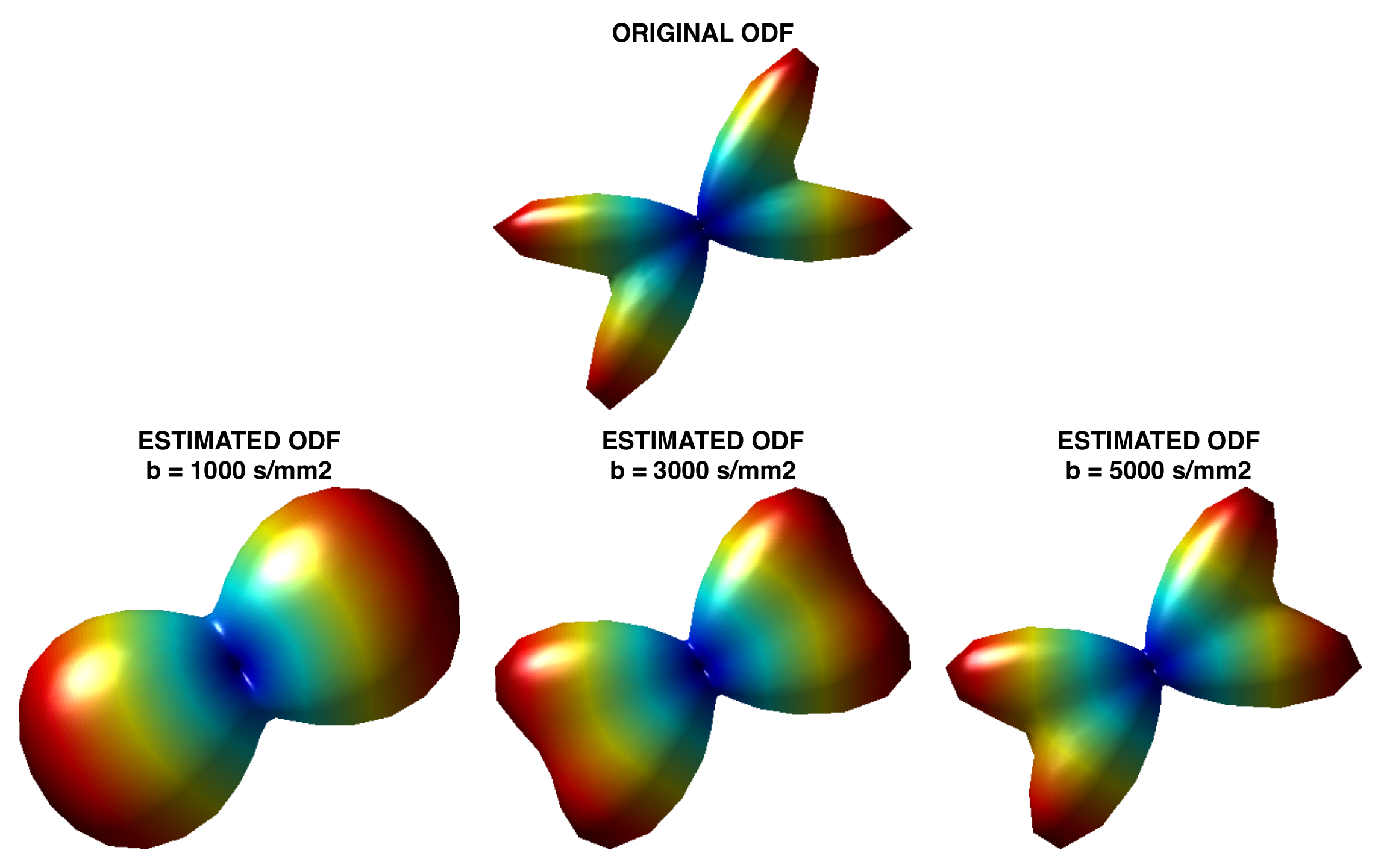} 
\caption{(Top) Original ODF; (Bottom) Estimated ODFs obtained from HARDI data generated with (left to right) $b$=1000, 3000, and 5000 s/mm$^2$.}
\label{F1}
\end{figure}

The nature of single-shell acquisition inherent in HARDI imposes constraints on the directional resolution of estimated ODFs, with higher values of the diffusion scintillation parameter $b$ (the so-called $b$-value) leading to better resolvability between various diffusion modes within a given voxel. This fact is exemplified in Fig.~\ref{F1}, which shows a simulated ODF\footnote{The ODF was generated using a standard Gaussian mixture model \cite{Alexander:2005fk} with equal volume fractions, FA=0.8 and MD = $7 \cdot 10^{-4}$ mm$^2$/s.} (top subplot) corresponding to two fibre tracts crossing each other at an angle of $60$$^{\rm o}$. At the same time, the bottom row of subplots depict the ODFs which have been recovered from the associated HARDI data generated with $b \in \{1000, 3000, 5000\}$ s/mm$^2$. One can see that the best directional resolution is attained at the maximum value of $b$= 5000 s/mm$^2$, as expected. It is also worthwhile noting that the above effect is intrinsic in both the Funk-Radon transform (FRT) \cite{Tuch:2004cr} and solid angle \cite{Aganj:2010fk,Tristan-Vega:2009uq} formulations of QBI.

As tempting as it might seem at the first glance, working with relatively high $b$-values is usually avoided in practice for two main reasons. First, an increase in the $b$-value is typically achieved through using longer mixing times, which unavoidably leads to substantially low values of signal-to-noise ratio (SNR) \cite{Bihan:2006th}. Second, using higher $b$-values makes the diffusion data less sensitive to the effects of fast diffusion, which is often associated with the concept of ``free water" \cite{Pasternak:2009qy}, and hence represents a diagnostically important diffusion regime. As a result, diffusion data are commonly acquired with relatively low values of $b$, typically around $1000$ s/mm$^2$. Needless to add, in such cases, poor directional resolution of ODFs may become an issue, especially for such applications as fibre tractography \cite{Malcolm:2009pt}.

The problem of limited directional resolution of QBI can be addressed using the framework of spherical deconvolution (SD) \cite{Healy:1998aa}. In this formulation, SD is used to recover a {\em fibre ODF} (fODF) which, as opposed to ODF, quantifies the likelihood of a neural fibre to have a (local) tangent vector of a certain orientation. The idea of using SD as a method for improving the directional resolution of QBI was first introduced in \cite{Tournier:2004tw}. In this initial approach, fODFs were recovered through a direct matrix inversion (aka naive deconvolution) preceded by low-pass filtering. A more advanced SD technique based on a least-square (LS) formulation with positivity constraints was proposed by the same authors in \cite{Tournier:2007ly, Tournier:2008ve}. An interesting extension to the later method was also discussed in \cite{Jeurissen:2011aa}, where the uncertainty in fODF estimation was reduced through a bootstrap procedure. In \cite{Anderson:2005kx}, the fOFD is parameterized based on a two-compartment model, followed by its reconstruction using non-regularized (naive) inverse filtering. A similar line of arguments is used in \cite{Sakaie:2007nx}, albeit this time with Tikhonov regularization implemented implicitly through damped SVD-based inversion. Moreover, as opposed to many other works in the field, this study addressed the important problem of automatic determination of an optimal value of the regularization parameter based on generalized cross validation. Advanced statistical considerations were also employed in \cite{Alexander:2005zr} to derive a maximum entropy deconvolution algorithm. The same idea of entropy maximization was later adopted by \cite{Jing:2012vn}. In the latter work, however, the reconstruction was performed within the framework of blind source separation, which offers the substantial advantage of independence on the knowledge of single fibre response. Unfortunately, the Infomax approach used by the authors in \cite{Jing:2012vn} does not explicitly take into account the distribution and level of measurement noises, which makes it overly dependent on measurement conditions. Finally, we would also like to mention the SD approach of \cite{Jian:2007il}, which is close in philosophy to the methods discussed in the present paper. In particular, \cite{Jian:2007il} advocates the idea of {\em sparse} SD, congruent with the assumption on an fODF to have a relatively small number of ``sharp" maxima in the directions of associated neural bundles.   

While different in their assumptions, computational requirements, and performances, virtually all the above-mentioned SD methods share a common drawback. Particularly, all these methods strongly adhere to the assumption on the data signal to be formed as a convolution of a {\it single-fibre response} (SRF) and an fODF, thereby explicitly requiring the presence of a fibre tract at the corresponding voxel. However, situations are frequent in which the data may alternatively be associated with either grey matter or cerebrospinal fluid (CSF), as well as with partial volume fraction voxels containing a mixture of different types of cerebral tissue. In such cases, the diffusion signals should be expected to have a non-negligible {\em isotropic} component, which comes at odds with the assumption on fODFs to be, e.g., sparse \cite{Jian:2007il} or of maximal entropy \cite{Alexander:2005zr}. 

The above-mentioned deficiency of earlier application of SD to dMRI has been alleviated in a series of more recent works on the subject. Thus, in \cite{Dellacqua:2007aa, Dellacqua:2010uq}, fODF has been supplanted by a different quantity, called a ``fibre orientation function" (FOF), which explains a combined effect of both (multimodal) anisotropic and isotropic diffusions. Subsequently, a variation of the Lucy-Richardson deconvolution algorithm was employed to recover the fODF, while gradually suppressing the isotropic component of the signal. The deconvolution algorithm in \cite{Schultz:2008ys} iteratively estimates and subtracts the isotropic component from HARDI data before fODFs are estimated. A data formation model similar to that of \cite{Dellacqua:2007aa, Dellacqua:2010uq} was also exployed by the authors of \cite{Yap:2012xx}, in combination with the sparse deconvolution ideas of \cite{Jian:2007il}. Quite a different type of deconvolution methods were explored in \cite{Kaden:2008xz, Kaden:2012aa} based on Bayesian inference. Specifically, \cite{Kaden:2012aa} takes advantage of a measure-theoretic framework to represent the fODF as a probability measure decomposable into three unique components, which allow adequate description of a wide spectrum of possible fODF shapes and patterns. Similarly to the previously mentioned results, the models used in \cite{Kaden:2008xz, Kaden:2012aa} explicitly account for the presence of isotropic diffusion.  

Since neutral bundles extend continuously in space, it is reasonable to assume spatially adjacent fODFs to exhibit a fair amount of correlation, which can be exploited to improve the results of SD. This idea has been exploited in several studies to improve the accuracy of QBI \cite{Goh:2009uq, Raj:2011fk}. In application to SD, spatial regularization was used in  \cite{Ramirez-Manzanares:2007fb}, in which case a spatial continuity of fODFs was enforced via minimizing a weighted quadratic penalty.  A more recent work in \cite{Reisert:2011ss} has introduced an anisotropic regularization scheme which, for any spatial direction $\uu$, minimizes the $\mathcal{L}_2$-norm of the projection of the spatial gradient of fODFs onto $\uu$. It has been shown that this minimization favours the spatial continuity of neural fibres, thereby producing anatomically plausible reconstructions. Unfortunately, neither of the above-mentioned approaches was designed to deconvolve HARDI signals in the presence of isotropic diffusion.

Despite the apparent success of SD in application to diffusion imaging, there is still much space for further improvements. In particular, it appears that little has been done on the development of SD algorithms which can perform reliably in the presence of isotropic diffusion, while imposing effective regularization constraints on {\em both} the isotropic and anisotropic components of diffusion signals. Needless to add, contriving such a reconstruction problem is likely to result in composite optimization, solving which could be a non-trivial problem by itself. Accordingly, the present paper aims to contribute to the existing body of works on SD in the following directions:
\begin{enumerate}   
\item We formulate a new method for SD of HARDI data, subject to spatial regularization of {\em both} the isotropic and anisotropic components of HARDI signals as well as their related ODFs/fODFs.
\item We describe a computationally efficient implementation of the proposed algorithm based on the idea of variable splitting \cite{Combettes:2011ee}. The proposed computational solution has a particularly simple modular structure, which is straightforward to reproduce using standard computational means.
\end{enumerate}

The remainder of the paper is organized as follows. Section~\ref{S2} details the data formation model used in the subsequent derivations of the proposed deconvolution approach, which is introduced in Section~\ref{S3}. Section~\ref{S4} provides details on our numerical implementation of the proposed method, while Section~\ref{S5} sets out the experimental setup used for its numerical validation. The results of application of the proposed method to both computer simulated and {\it in vivo} data are reported in Section~\ref{S6}. The section also contains a comparative analysis of the performance of the proposed algorithm against that of several alternative solutions. Finally, Section~\ref{S7} finalizes the paper with recapitulation of its results and conclusions.

\section{Data formation model}\label{S2}
To fix the ideas, we start with a formal setting, in which HARDI data are assumed to be collected over a bounded (open) subset of $\Omega \in \Artri$. In this case, for each $\rr \in \Omega$, a HARDI signal $s(\uu | \rr)$ can be viewed as a positive-valued, spherical function $s(\cdot | \rr): \Estu \to [0, \infty)$, with $\Estu$ denoting the unit sphere in $\Artri$, in which case $\uu \in \Estu$ is interpreted as the direction of diffusion encoding. When a given voxel of interest supports a single neural fibre, the corresponding EAP can be closely approximated by a unimodal Gaussian density, in which case the HARDI signal $s(\uu | \rr)$ can be described as \cite{Basser:1994pd, Bihan:1986lp}
\begin{equation}\label{DTI}
s(\uu | \rr) = s_0(\rr) \exp \{ - b~\uu^T D(\rr) \uu \}, \quad \forall \rr \in \Omega,
\end{equation}
where the diffusion tensor $D(\rr)$ encodes the directivity and ellipticity of the pattern of local diffusion, while the $b$-value is typically set in the range between 1000 and 3000 s/mm$^2$. Note that the $b_0$-image $s_0(\rr)$ is usually acquired through additional measurements and used to normalize the HARDI signals. For the sake of notational convenience, in what follows, the signal $s(\uu | \rr)$ will be assumed to be normalized, implying $s_0(\rr) = 1$, for all $\rr \in \Omega$.

Although standard in DTI, the model of \eqref{DTI} is not applicable in situations when a voxel of interest supports multiple neural bundles \cite{Tuch:2002nx}. If this is the case, then under some fairly general conditions \cite{Alexander:2005fk}, the HARDI signal $s(\uu | \rr)$ can be assumed to be formed as a (linear) superposition of several ``DTI signals" weighted by their respective partial volume fraction coefficients. To formalize such a signal formation model, it is common to use the notion of an SFR \cite{Tournier:2004tw}, which is, in fact,  equal to the elementary ``DTI signal" given by \eqref{DTI}. Moreover, in the SD literature, for the definition of a SFR $h(\uu)$ it is standard to use a cylindrically symmetric diffusion tensor $D_0 = {\rm diag} \{ \LMIN,~\LMIN,~\LMAX\}$ (with $\LMAX > \LMIN$), which seems to naturally comply with an expected (local) geometry of neural fasciculi. In such a case, it is straightforward to show that the SFR becomes a {\em zonal} (spherical) function \cite{Freeden:1998ud} that is formally given by
\begin{equation}\label{SFR} 
h_{\vv_0}(\uu) = h(\uu \cdot \vv_0) = \alpha \exp\{-\beta (\uu \cdot \vv_0) \},
\end{equation}
with the dot standing for the standard (Euclidean) dot product in $\Artri$, $\alpha \triangleq \exp\{ - b \LMIN\}$, $\beta \triangleq b (\LMAX-\LMIN)$, and $\vv_0 = [0,~0,~1]^T$ denoting the north pole of $\Estu$.

It is worthwhile noting that, due to the property of the SFR in \eqref{SFR} to be zonal, its value at a given $\uu$ depends only on the angle between $\uu$ and a fixed direction $\vv$ (e.g., $\vv = \vv_0$, as in \eqref{SFR}), which makes $h_\vv(\uu)$ invariant under rotations around $\vv$. This invariance allows the result of convolution with $h_\vv(\uu)$ to be expressed as a function of $\Estu$ (rather than of the orthogonal group $SO(3)$ \cite{Healy:1998aa}), in which case our signal formation model becomes 
\begin{equation}\label{MOD1}
s(\uu | \rr) = \int_{\Estu} h_\uu(\vv) d \mu(\vv | \rr) = \int_{\Estu} h(\uu \cdot \vv) d \mu(\vv | \rr), \quad \forall \rr \in \Omega. 
\end{equation}
Here $\mu(\uu | \rr)$ is a probability measure that is used to model the fibre probability distribution over $\Estu$ \cite{Kaden:2012aa}. In particular, at any $\rr \in \Omega$, $\mu (\cdot | \rr): \mathcal{B} \to [0, \infty]$ quantifies the relative frequency of specific fibre orientations over a given element of the Borel sigma algebra $\mathcal{B}$ of $\Estu$.

A mathematically elegant and physiologically meaningful way to interpret the structure of $\mu$ was recently described in \cite{Kaden:2012aa}, where the authors took advantage of the Lebesgue's decomposition theorem to represent $\mu$ as a sum of three components ({\it viz.}, discrete, absolutely continuous, and singular continuous), each of which is able to model a distinct characteristic of the fibre orientation distribution. In the current paper, however, we proceed under a simplified assumption on $\mu$ to be absolutely continuous, in which case it can be described in terms of a non-negative, Borel measurable function $f(\uu | \rr)$ as $d \mu(\vv | \rr) = f(\vv | \rr) d \eta(\vv)$, with $\eta$ being the Haar measure of $\Estu$. We note that this simplification seems to be reasonable, considering the fact that, in practical computations, both measurements and resulting estimates are always bounded in value and discrete. 

The above simplifying assumption leads to the standard (forward) model for $s(\uu | \rr)$ which reads \cite{Tournier:2008ve, Jian:2007il, Dellacqua:2010uq, Kaden:2012aa, Yap:2012xx}
\begin{equation}\label{MOD2}
s(\uu | \rr) = \int_{\Estu} h(\uu \cdot \vv)  f(\vv | \rr) d \eta(\vv), \quad \forall \rr \in  \Omega,
\end{equation}
in which case the density $f(\uu | \rr)$ is conventionally referred to as a {\em fibre orientation distribution function} (fODF). It should also be noted that the models in \eqref{MOD1} and \eqref{MOD2} are {\em stationary}, since the SFR is assumed to be fixed within a given voxel as well as across the whole image domain $\Omega$. While only approximative \cite{Parker:2013ee}, the stationary SD model has nevertheless shown to yield useful reconstructions, while offering the important advantage of tractability and amenability to numerical computations.     

Additionally, the structure of fODF $f(\uu | \rr)$ in \eqref{MOD2} deserves a special consideration. Since the measured diffusion signal receives contributions from both coherently ordered axonal fascicles as well as from their complex and more heterogeneous extra-axonal surroundings (containing astrocytes, glia, and randomly oriented extracellular matrix molecules) \cite{Assaf:2004kx}, it seems reasonable to consider $f(\uu | \rr)$ to be composed of two main terms, {\it viz.} anisotropic and isotropic. Specifically, following the line of ideas advocated in \cite{Dellacqua:2010uq, Kaden:2012aa, Yap:2012xx}, we model $f(\uu | \rr)$ according to
\begin{equation}\label{DECOMP}
f(\uu | \rr) = (1 - p_{\rm iso}(\rr)) f_a(\uu | \rr) + p_{\rm iso}(\rr) f_{\rm iso}(\rr), 
\end{equation}
with $f_a(\uu | \rr)$ and $f_{\rm iso}(\rr)$ representing the anisotropic and isotropic components of the fibre probability distribution, respectively, and $0 \le p_{\rm iso}(\rr) \le 1$ controlling their partial volume fractions at $\rr \in \Omega$. Moreover, $f_{\rm iso}(\rr)$ can be further represented as $f_{\rm iso}(\rr) = \exp(-b \lambda_{\rm iso}(\rr))$, where $\lambda_{\rm iso}(\rr) > 0$ is the apparent diffusivity of the isotropic component \cite{Kaden:2012aa, Yap:2012xx}.
 
Since the isotropic components $f_{\rm iso}(\rr)$ in \eqref{DECOMP} is dissociated from the anisotropic component $f_a(\cdot | \rr)$, it would no longer be correct to regard $f(\uu | \rr)$ as an fODF. To overcome this notational inconsistence, from now on, we will use this term when referring to $f_a(\uu | \rr)$ instead. At the same time, since $f_{\rm iso}(\rr)$ can be viewed as a scalar-valued function of $\rr \in \Omega$, it will be referred hereinafter to as an {\em isotropic diffusion map} (IDM).   

Finally, in this paper, instead of trying to recover $f_a(\uu | \rr)$ and $f_{\rm iso}(\rr)$ along with $p_{\rm iso}(\rr)$, we estimate the weighted quantities $f_a^\prime(\uu | \rr) \triangleq (1 - p_{\rm iso}(\rr)) f_a(\uu | \rr)$ and $f_{\rm iso}^\prime(\rr) \triangleq p_{\rm iso}(\rr) f_{\rm iso}(\rr)$, such that $f(\uu | \rr) = f_a^\prime(\uu | \rr) + f_{\rm iso}^\prime(\rr)$, $\forall \rr$. We note that, when normalized by $\int_\Estu f(\uu | \rr) d \eta(\uu)$, $f_a^\prime(\uu | \rr)$ and $f_{\rm iso}^\prime(\rr)$ acquire the ``flavour" of posterior probabilities. Thus, for example, the values of $f_a^\prime(\uu | \rr)$ are not only indicative of the orientations of neural fibres, but also reflect one's {\em level of confidence} that the fibres are {\em actually} present at a given location $\rr$ at the first place. We support this concept through an experimental study and argue that using the weighted densities $f_a^\prime(\uu | \rr)$ and $f_{\rm iso}^\prime(\rr)$ can benefit a number of related applications, such as probabilistic fibre tractography \cite{Malcolm:2009pt}.

\section{Proposed approach}\label{S3}
\subsection{Model discretization}
As usual, the formalism of matrix-vector multiplications turns out to be the most convenient for formulation of practical solutions. To this end, we first note that (normalized) HARDI data typically consists of a set of diffusion-weighted scans acquired for $K$ directions $\{\uu_k\}_{k=1}^K$ of diffusion-encoding gradients (with $\uu_k \in \Estu, \forall k$). For the convenience of exposition, we concatenate these scans into $K$ {\em row} vectors of length $I$, with $I$ being equal to the number of spatial samples (voxels) within $\Omega$. Further, the vectors thus obtained can be organized as the rows of a $K\times I$ matrix $s$, in which case the columns of $s$ correspond to diffusion measurements observed at different spatial locations. In what follows, we use both super- and subscripts to distinguish between the columns $\{s^i\}_{i=1}^I$ and rows $\{s_k\}_{k=1}^K$ of the data matrix $s$.

Discretizing the SD model of \eqref{MOD2} is the next step. To this end, let $\{\vv_j\}_{j=1}^J$ (with $J > K$) be a set of spherical points over which the values of the fODFs are to be recovered. Then, given estimates\footnote{To estimate these parameters, it is standard to fit the DTI model of \eqref{DTI} to HARDI signals corresponding to, e.g., corpus callosum. As the latter is predominantly composed of single (commisural) fibre bundles, its related signals can provide a reliable estimate of the SFR, subject to appropriate averaging \cite{Tournier:2004tw}.} $\tilde{\alpha}$  and $\tilde{\beta}$ of the SFR parameters in \eqref{SFR}, we define the $(k,j)$-th element of a $K\times J$ matrix $H = \{ h_{k,j} \}$ according to 
\begin{equation}\label{Hmatrix}
h_{k,j} = \tilde{\alpha} \exp\{ -\tilde{\beta} (\uu_k \cdot \vv_j) \}.
\end{equation}
Note that the $J$ columns of resulting $H$ correspond to the SFR rotated in directions $\{\vv_j\}_{j=1}^J$ and discretized at points $\{\uu_k\}_{k=1}^K$. Finally, we define $\Phi$ to be a $K \times (J+1)$ matrix obtained from $H$ through addition of an extra column of ones, {\it viz.}
\begin{equation}\label{Pmatrix}
\Phi = [H ~~ {\bf 1}],
\end{equation}
where ${\bf 1} = [1, 1, ..., 1]^T \in \mathbb{R}^K$.

Now, let $\rr_i$ be the coordinate of an arbitrary voxel within $\Omega$, with $i=1,2,\ldots,I$, and let $f^i \in \mathbb{R}^{J+1}$ be a column vector defined as $f^i = [f_a^\prime(\vv_1 | \rr_i), f_a^\prime(\vv_2 | \rr_i), \ldots, f_a^\prime(\vv_J | \rr_i), f_{\rm iso}^\prime(\rr_i) ]^T$. Then, in the absence of measurement noises and disregarding the effect of discretization, the SD model \eqref{MOD2} along with \eqref{DECOMP} suggest that $s^i = \Phi f^i$, for all $i = 1, 2, \ldots, I$. It is, therefore, convenient to agglomerate all the above model equations into a single one that reads
\begin{equation}\label{MOD3}
s = \Phi f, 
\end{equation} 
where $f$ is a $(J+1)\times I$ matrix, with its columns defined by $f^i$, with $i = 1, 2, \ldots, I$. Note that the last row $f_{J+1}$ of $f$ is equal to a row-stacked version of the IDM, whilst the first $J$ rows $f_1, \ldots, f_{J}$ of $f$ can be viewed as row-stacked versions of the images obtained by restricting the fODF $f_a(\uu|\rr)$ to the directions $\vv_1, \ldots, \vv_J$, respectively. 

\subsection{Estimation framework} For obvious reasons, recovering a useful estimate of $f$ based on model \eqref{MOD3} alone is a futile exercise. According to the formalism of Bayesian estimation, to render the reconstruction unique and stable, the model equation needs to be augmented with reasonable {\it a priori} assumptions on the nature of $f$ in \eqref{MOD3}. One of such assumptions, which has been proved to be particularly useful for reconstruction of fODFs, is that of {\em sparsity}. Indeed, the anatomical organization of white matter suggests the number of axonal fascicles running through any given voxel $\rr_i$ is likely to be relatively small. This fact, in turn, implies that the vector $[f_a^\prime(\vv_1 | \rr_i), f_a^\prime(\vv_2 | \rr_i), \ldots, f_a^\prime(\vv_J | \rr_i) ]^T$ could be reasonably expected to have a relatively small number of significant components, with the rest of its entries distributed in a close proximity of zero. Such a behaviour of $f_a(\uu|\rr)$ can be modelled in a number of different ways \cite{Jian:2007il, Kaden:2012aa,Yap:2012xx}. In the present paper, we take advantage of the standard method of recovering a sparse vector through minimization of its $\ell_1$-norm \cite{Jian:2007il}.

Before proceeding to the next step, we note that the SD model of \eqref{MOD3} admits an alternative interpretation, according to which every $s^i$ is approximated by a linear combination of the columns of $\Phi$. As the last (constant) column of $\Phi$ has distinctly different {\em morphological} properties as compared to the other columns of the matrix, it seems reasonable to minimize the $\ell_1$-norm of the {\em entire} vector $f^i$ (rather than only of its part associated with $f_a^\prime(\uu|\rr_i)$). In this case, the parsimonious nature of $\ell_1$ minimization will force the optimal solution to be dominated by either an IDM or an fODF component, while permitting them both only when there is clear evidence of their concurrent existence. It deserves noting that a similar principle has been used in morphological component analysis to decompose a signal of interest into morphologically distinct components \cite{Michailovich:2002fe, Starck:2004kc}.

To avoid unnecessary complications in notations, in the derivations that follow, we use $\| \cdot \|_2$ and $\| \cdot \|_1$ to denote the $\ell_2$- and $\ell_1$-norms of vectors as well as the analogous ``entry-wise" norms of matrices. Thus, for example, the $\ell_2$-norm of a HARDI signal $s$ can be expressed in two ways as $\| s \|_2^2 = \sum_{i=1}^I \| s^i \|_2^2  = \sum_{k=1}^K \| s_k \|_2^2$, while the $\ell_1$-norm of $f$ can be expressed as $\| f \|_1 = \sum_{i=1}^I \| f^i \|_1  = \sum_{j=1}^{J+1} \| f_j \|_1$. With this notation at hand, we formulate the problem of finding an optimal $f^\ast$ as
\begin{align}\label{L1}
f^\ast &= \arg \min_f \| f \|_1  \\
\mbox{subject to} \,\, &\| \Phi f - s \|_2^2 \le \epsilon, \,\, f \ge 0 \notag,
\end{align}
where $\epsilon$ controls the size of measurement and model errors, whereas the point-wise inequality constraint is added to assure that both the IDM and fODFs are non-negative quantities. Note that the problem in \eqref{L1} is analogous to the one described in \cite{Yap:2012xx}, apart from the fact that \eqref{L1} applies to a whole set of HARDI data, rather than to a single voxel.

To facilitate numerical solution, the problem \eqref{L1} is usually reformulated in its equivalent, unconstrained (Lagrangian) form as given by   
\begin{equation}\label{L1L}
f^\ast = \arg \min_f \left\{ \frac{1}{2} \| \Phi f - s \|_2^2 + \lambda \| f \|_1 + \varphi_{\ge}(f) \right\}, 
\end{equation}
where $\lambda > 0$ is a user-controlled regularization parameter and $\varphi_\ge$ denotes the indicator function of the positive orthant. Specifically, $\varphi_\ge(f) = 0$, if {\em all} entries of $f$ are non-negative, and $\varphi_\ge(f) = +\infty$, otherwise. The problem \eqref{L1L} can be solved using a variety of methods of non-smooth optimization (e.g., \cite{Daubechies:2003bv}). 

Unfortunately, the solution of \eqref{L1L} could be only suboptimal, as it completely disregards any spatial-domain dependencies between the values of $f$. To palliate this deficiency, we first note that each {\em row} of $f$ can be considered to be a discrete image defined over the spatial lattice $\{\rr_i\}_{i=1}^I \in \Omega$ and stacked into a row vector. Particularly, in this interpretation, the first $J$ rows $\{f_j\}_{j=1}^J$ can be viewed as {\em restrictions} of the fODF to directions $\{\vv_j\}_{j=1}^J$, while $f_{J+1}$ represents the corresponding IDM. Naturally, the images $\{f_j\}_{j=1}^J$ and $f_{J+1}$ have different statistical properties, and therefore they should be regularized in different ways. Thus, to spatially regularize the fODF $f_a(\uu|\rr)$ we adopt the fibre continuity approach of \cite{Reisert:2011ss}, which requires the directional derivative of $f_j$ along $\vv_j$ to be relatively small in value. Formally, let $\nabla_d: \mathbb{R}^{I} \to \mathbb{R}^{I}$, with $d=1,2,3$, denote the operators of spatial differencing in the direction of $x$, $y$, and $z$ coordinates, respectively. Then, for each $j = 1, \ldots, J$, one can then assemble a $3 \times I$ matrix $D f_j$, with its rows defined by the partial differences of $f_j$, {\it viz.} $D f_j = [ \nabla_1 f_j^T ~ \nabla_2 f_j^T ~ \nabla_3 f_j^T ]^T \in \mathbb{R}^{3\times I}$. Subsequently, with $\vv_j$ being a column vector, the directional derivative of $f_j$ along $\vv_j$ is conveniently given by $\vv_j^T D f_j$, in which case the approach of \cite{Reisert:2011ss} calls for minimizing $\sum_{j=1}^J \| \vv_j^T D f_j \|_2^2$.    

It should be noted that the above approach cannot be extended to the IDM $f_{J+1}$, since the latter is devoid of directional continuity \cite{Reisert:2011ss}. Yet, for the pure sake of harmonizing the notations, we replace the original minimization of $\sum_{j=1}^J \| \vv_j^T D f_j \|_2^2$ by minimizing $\sum_{j=1}^{J+1}  (1-\delta_{j,J+1})  \| \vv_j^T D f_j \|_2^2$, where $\delta_{t,r}$ stands for the Kronecker symbol which obeys $\delta_{t,r} = 1$, if $t=r$, while $\delta_{t,r} = 0$, otherwise. (Note that, since the last component in the above summation is multiplied by a zero weight, the choice of $\vv_{J+1}$ is immaterial and it has no effect on the reconstruction procedure as shown later in the paper.) Subsequently, the optimal solution can still be defined as a global minimizer over $f$, which is now given by                                     
\begin{equation}\label{L1A}
f^\ast = \arg \min_f \left\{ \frac{1}{2} \| \Phi f - s \|_2^2 + \lambda \| f \|_1 + \mu \| f \|_a^2 + \varphi_{\ge}(f) \right\}, 
\end{equation}
where
\begin{equation}\label{Anorm}
\| f \|_a^2 = \sum_{j=1}^{J+1}  (1-\delta_{j,J+1})  \| \vv_j^T D f_j \|_2^2,
\end{equation}
and $\mu > 0$ is another regularization constant.

Finally, the problem of regularizing the IDM $f_{J+1}$ must not be overseen as well. Since one can reasonably expect $f_{J+1}$ to vary smoothly throughout the brain with the exception of abrupt changes between white matter, grey matter, CSF, as well as the regions of possible brain pathologies, it seems justified to model $f_{J+1}$ as a function of bounded variation (BV) \cite{Rudin:1992fh}. In particular, the BV model enforces the assumption on $f_{J+1}$ to have a relatively small value of its {\em total variation} (TV) seminorm $\| f_{J+1} \|_{TV}$ which can be defined as follows. Let $\nabla_1 f_{J+1}[i]$, $\nabla_2 f_{J+1}[i]$, and $\nabla_3 f_{J+1}[i]$ denote the $i$-th elements of vectors $\nabla_1 f_{J+1}$, $\nabla_2 f_{J+1}$, and $\nabla_3 f_{J+1}$, respectively. Then, the TV seminorm of $f_{J+1}$ can be defined in a standard way as
\begin{equation}\label{TV}
\| f_{J+1} \|_{TV} =  \sum_{i=1}^I \Big[ \sum_{d=1}^3 | \nabla_d f_{J+1}[i] |^2 \Big]^{1/2}.
\end{equation}

It is definitely possible to apply the definition of TV to images $f_j$, with $j = 1, \ldots, J$ as well. However, minimizing these norms would likely mislead the estimation process, since $f_j$ may not be assumed to be piecewise smooth, in general. Still, to balance the notations, we define the TV seminorm of $f$ according to
\begin{equation}\label{TVf}
\| f \|_{TV} = \sum_{j=1}^{J+1} \delta_{j,J+1} \| f_j \|_{TV},
\end{equation}
which leads to the optimal solution of the form
\begin{equation}\label{Cost}
f^\ast = \arg \min_f \left\{ \frac{1}{2} \| \Phi f - s \|_2^2 + \lambda \| f \|_1 + \mu \| f \|_a^2 + \nu \| f \|_{TV} + \varphi_{\ge}(f) \right\}, 
\end{equation}
with $\nu >0$ being an additional regularization constant which controls the piecewise smooth behaviour of the IDM. We admit that automatically determining an optimal value of $\nu$, as well as those of $\lambda$ and $\mu$, is a difficult problem, which extends well beyond the scope of the current paper. It was observed in practice, however, that finding acceptable values of these parameters by wonted trials-and-errors is a much less arduous task, as it might seem at the first glance.  

The solution of \eqref{Cost} entails minimizing a non-smooth cost function, which effectively rules out the use of gradient-based methods of numerical optimization. Moreover, the composite nature of the cost makes it difficult to devise an efficient optimization approach which would perform the minimization {\em directly} with respect to $f$. To overcome these difficulties, the next section introduces a particularly simple solution using the alternating directions method of multipliers (ADMM) \cite{Boyd:2010kx}. Apart from breaking down the optimization in \eqref{Cost} into a sequence of simple and closed-form solutions, the method offers a straightforward approach to splitting the computations between multiple computing cores/units, which is a significant advantage considering the relatively large dimensionality of HARDI data. 

\section{Numerical solution}\label{S4}
To simplify the solution of \eqref{Cost}, we introduce two auxiliary variables $u$ and $v$, and replace the original optimization problem by an equivalent, equality-constrained one. Specifically, 
\begin{align}\label{CC}
\min_{f, u, v} \Big\{ \frac{1}{2} \| \Phi f &- s \|_2^2 + \lambda \| u \|_1 + \mu \| v \|_a^2 + \nu \| v \|_{TV} + \varphi_{\ge}(u) \Big\}, \\
&\mbox{subject to} \,\, f = u, \,\, f = v \notag
\end{align}
Note that in \eqref{CC} the minimization is carried out with respect to three variables, namely $f$, $u$, and $v$. A standard approach to solving such equality-constrained problems is based of the use of augmented Lagrangian methods \cite{Boyd:2004fy}. Particularly, for the case at hand, this approach amounts to the following iterations (starting from some intitial values of Lagrange multipliers $p_u^0$ and $p_v^0$, e.g., $p_u^0 = p_v^0 = 0$). 
\begin{align}\label{AUG}
(f^{(t+1)}, u^{(t+1)}, v^{(t+1)}) = \arg \min_{f, u, v} \Big\{ \frac{1}{2} \| \Phi f - s \|_2^2 + \lambda \| u \|_1 + \mu \| v \|_a^2  + \nu \| v \|_{TV} + \notag \\
+ \varphi_{\ge}(u) + \frac{\delta_u}{2} \| f - u + p_u^{(t)} \|_2^2 +  \frac{\delta_v}{2} \| f - v + p_v^{(t)} \|_2^2 \Big\} \\
p_u^{(t+1)} = p_u^{(t)} + f^{(t+1)} - u^{(t+1)}  \hspace{3in} \notag \\
p_v^{(t+1)} = p_v^{(t)} + f^{(t+1)} - v^{(t+1)}  \hspace{3in} \notag
\end{align}
where $t$ stands for an iteration index, while $\delta_u >0$ and $\delta_v>0$ are some positive constants\footnote{Note that the algorithm is guaranteed to converge for any positive $\delta_u$ and $\delta_v$. In the present work we use $\delta_u$ = $\delta_v$ = 0.5.}.

Finally, in ADMM, the concurrent minimization with respect to $f$, $u$, and $v$ is replaced by sequential minimization with respect to $f$, $u$, and $v$ independently. In this case, starting from some $f^{(0)} = u^{(0)} = v^{(0)}$, the update proceeds according to:
\begin{align}
&{\bf Step 1:} \,\, f^{(t+1)} = \arg \min_f \Big\{ \frac{1}{2} \| \Phi f - s \|_2^2 + \frac{\delta_u}{2} \| f - u^{(t)} + p_u^{(t)} \|_2^2 +  \frac{\delta_v}{2} \| f - v^{(t)} + p_v^{(t)} \|_2^2 \Big\} \label{ST1} \\
&{\bf Step 2:} \,\, u^{(t+1)} = \arg \min_u \Big\{ \frac{\delta_u}{2} \| f^{(t+1)} - u + p_u^{(t)} \|_2^2  +  \lambda \| u \|_1 + \varphi_{\ge}(u) \Big\} \label{ST2} \\
&{\bf Step 3:} \,\, v^{(t+1)} = \arg \min_v \Big\{ \frac{\delta_v}{2} \| f^{(t+1)} - v + p_v^{(t)} \|_2^2  + \mu \| v \|_a^2  + \nu \| v \|_{TV} \Big\} \label{ST3},
\end{align} 
followed by the ``dual" step of updating the Lagrange multipliers $p_u$ and $p_v$, as specified in \eqref{AUG}. Although the ADMM approach has effectively supplanted a single minimization problem \eqref{Cost} by three minimization problems, the latter admit much simpler solutions as detailed below.

\subsection{Solution to Step 1}
The optimization problem in \eqref{ST1} is a classical least-square (LS) problem which admits a closed-form solution as given by
\begin{equation}\label{SOL1}
f^{(t+1)} = \left( \Phi^T \Phi + (\delta_u+\delta_v) I_{J+1} \right)^{-1} \big( \Phi^T s + \delta_u (u^{(t)} - p_u^{(t)}) + \delta_v (v^{(t)} - p_v^{(t)}) \big),
\end{equation}
where $I_{J+1}$ stands for a $(J+1) \times (J+1)$ identity matrix. To facilitate the computations, the inverse matrix $R \triangleq  \big( \Phi^T \Phi + (\delta_u+\delta_v) I_{J+1} \big)^{-1}$ can be precomputed and stored before the reconstruction procedure is initiated. 

\subsection{Solution to Step 2}
The optimization problem in \eqref{ST2} can be equivalently rewritten as
\begin{equation}
u^{(t+1)} = \arg \min_u \Big\{ \frac{1}{2} \| u - (f^{(t+1)}+ p_u^{(t)}) \|_2^2  +  \frac{\lambda}{\delta_u} \| u \|_1 + \varphi_{\ge}(u) \Big\}.
\end{equation}
In the absence of the indicator function $\varphi_\ge$, the solution to the above problem would be given by soft thresholding, {\em viz.} $u^{(t+1)} = \mathcal{S}_{\lambda/\delta_u} \{ f^{(t+1)}+ p_u^{(t)} \}$, with $\mathcal{S}_\tau (x) = {\rm sign}(x) (|x| - \tau)_+$ (where the subscript $+$ denotes the operation of keeping the positive part of the argument). As unexpected as it might sound, however, incorporating the indicator function actually simplifies the solution by replacing   the operator $\mathcal{S}_\tau (x)$ with its positively rectified version $\mathcal{S}_\tau^+ (x) = (x - \tau)_+$. Consequently, the solution to \eqref{ST2} is given by
\begin{equation}\label{SOL2}
u^{(t+1)} = \mathcal{S}_{\lambda/\delta_u}^+ \{ f^{(t+1)}+ p_u^{(t)} \} = \big( f^{(t+1)}+ p_u^{(t)} - \lambda/\delta_u \big)_+.
\end{equation} 

\subsection{Solution to Step 3} 
To derive an update equation for $v^{(t)}$, it is convenient to rewrite the optimization problem in \eqref{ST3} in a slightly different form as given by
\begin{equation}\label{VUP}
v^{(t+1)} = \arg \min_v \Big\{ \frac{1}{2} \| v - (f^{(t+1)}  + p_v^{(t)}) \|_2^2  + \frac{\mu}{\delta_v} \| v \|_a^2  + \frac{\nu}{\delta_v} \| v \|_{TV} \Big\}
\end{equation}
Furthermore, since in our (simplified) notations $\| v \|_2^2 = \sum_{j=1}^{J+1} \| v_j \|_2^2$, then using the definitions of $\| v\|_a^2$ and $\| v \|_{TV}$ in \eqref{Anorm} and \eqref{TVf}, correspondingly, one can reexpress \eqref{VUP} in an equivalent form as
\begin{align}\label{Vcost}
v^{(t+1)} = &\arg \min_v \Big\{ \sum_{j=1}^{J+1} \Big( \frac{1}{2} \| v_j - (f^{(t+1)}  + p_v^{(t)})_j \|_2^2  + \\
&+ \frac{\mu}{\delta_v} (1-\delta_{j,J+1})  \| \vv_j^T D v_j \|_2^2  + \frac{\nu}{\delta_v} \delta_{j,J+1} \| v_j \|_{TV}\Big) \Big\}, \notag
\end{align}
where $(f^{(t+1)}  + p_v^{(t)})_j$ denotes the $j$-th row of matrix $f^{(t+1)}  + p_v^{(t)}$.  

A closer look at \eqref{Vcost} reveals that its cost functional consists of $J+1$ positive-valued terms, each of which depends on the variables $v_j$, $j=1,2,\ldots,J+1$, independently. Hence, the minimization over $v$ can be replaced by  minimizations of the $J+1$ summands in \eqref{Vcost} with respect to their respective variables (i.e., $v_j$). In particular, for $j = 1, \ldots, J$, the resulting minimization problems are given by 
\begin{equation}\label{VJ}
v_j^{(t+1)} = \arg \min_{v_j} \Big\{ \frac{1}{2} \| v_j - (f^{(t+1)}  + p_v^{(t)})_j \|_2^2  + \frac{\mu}{\delta_v} \| \vv_j^T D v_j \|_2^2 \Big\}.
\end{equation}
The above is a simple LS problem, which can be solved using spectral methods (i.e., by means of linear filtering) as detailed in the Appendix. Note that the cost function in \eqref{VJ} does not contain a TV term due to mutual exclusivity of the weights in definitions of $\| v \|_a^2$ and $\| v \|_{TV}$. For the same reason, optimization over $v_{J+1}$ does not contain a fibre continuity term, resulting in
\begin{equation}\label{VTV}
v_{J+1}^{(t+1)} = \arg \min_{v_{J+1}} \Big\{ \Big( \frac{1}{2} \| v_{J+1} - (f^{(t+1)}  + p_v^{(t)})_{J+1} \|_2^2  + \frac{\nu}{\delta_v} \| v_{J+1} \|_{TV}\Big\}.
\end{equation}
Note that the problem in \eqref{VTV} is a classical TV-regularized denoising problem \cite{Rudin:1992fh}, which can be efficiently solved by a variety of different algorithms. In the current work, we used the semi-implicit, fixed-point approach of \cite{Chambolle:1997qo} due to its impressive numerical stability and fast convergence.

\section{Materials and Methods}\label{S5}
\subsection{Sources of data}\label{S5-1}
\begin{figure}[!t]
\centering
\includegraphics[width=5in]{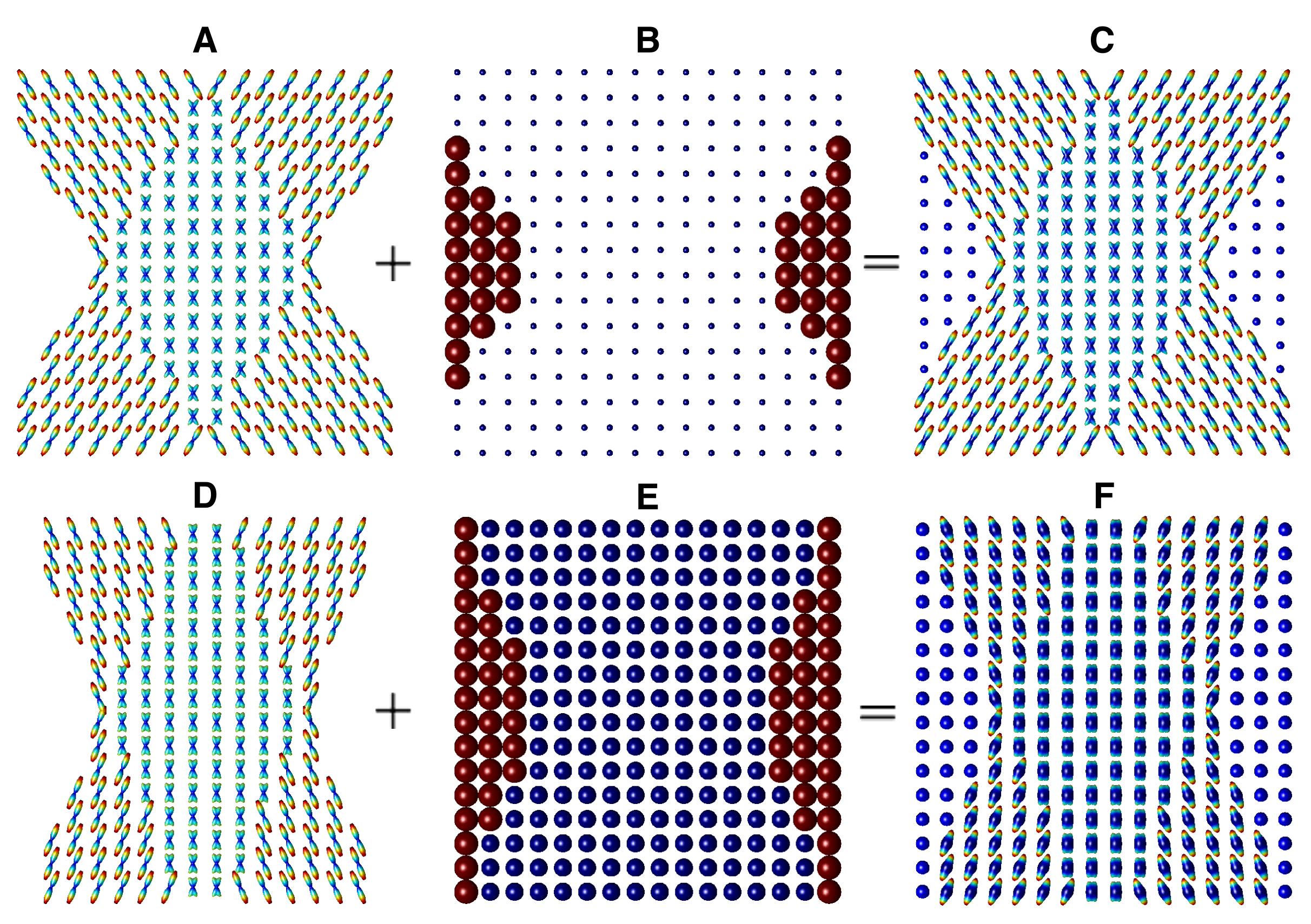} 
\caption{(Subplots A-C, left-to-right) The anisotropic, isotropic, and the combined field of ODFs associated with {\em in silico} HARDI signals generated with $b = 3000$ s/mm$^2$, $\alpha = 60^\circ$, and $p_{iso} = 0.25$; (Subplots D-F) Same as above, only with $\alpha = 45^\circ$ and $p_{iso} = 0.75$.}
\label{F2}
\end{figure}

The performance of the proposed and reference methods has been assessed using both computer-simulated and real-life data. Specifically, the simulated data were generated based on a standard Gaussian mixture model \cite{Alexander:2005fk} with FA=0.8 and MD = $7 \cdot 10^{-4}$ mm$^2$/s (which corresponds to $\LMAX=17 \cdot 10^{-4}$ mm$^2$/s and $\LMIN = 3 \cdot 10^{-4}$ mm$^2$/s). The spatial dimensions of the data were set to be equal to $16\times16\times12$, while the directions of diffusion encoding were defined by the 2nd order tessellation of icosahedron restricted to the northern hemisphere of $\Estu$ (thus resulting in $K=81$ sampling points). The data were designed so as to emulate a crossing of two cylindrically symmetric ``fibres" of 8 voxels in diameter, with the crossing angle $\alpha$ in the range $\alpha \in [30^\circ, 90^\circ]$ with step size $5^\circ$. Subplots A and D of Fig.~\ref{F2} depict the {\em theoretical} ODFs corresponding to the middle ``layer" of two signal arrays which have been synthesized for $\alpha=60^\circ$ and $\alpha=45^\circ$, respectively. Additionally, to allow investigation of the effect of isotropic diffusion on the accuracy of SD-based reconstruction, the simulated signals have also been combined with a constant (i.e., isotropic) component of a variable magnitude. The diffusivity of the isotropic component was set to be equal to $8 \cdot 10^{-4}$ mm$^2$/s, whereas its partial volume fractions outside and inside of the ``fibres" were set to 1 and $p_{iso}$, respectively, with $p_{iso} \in \{0, 0.25, 0.5, 0.75\}$. Two examples of the isotropic component are shown in Subplots B and E of Fig.~\ref{F2} for the case of $p_{iso}=0.25$ and $p_{iso}=0.75$, correspondingly. (Note that, in the above subplots, the {\em absolute} size of the glyphs has been optimized for visualization, and thus it does not represent the actual values of the isotropic ODFs.) Finally, Subplot C of Fig.~\ref{F2} depicts combined ODFs obtained as a result of the summation of the ODFs shown in Subplots A and B, while the result of the summation of the ODFs in Subplots D and E is displayed in Subplot F of the same figure.

\begin{figure}[!t]
\centering
\includegraphics[width=6in]{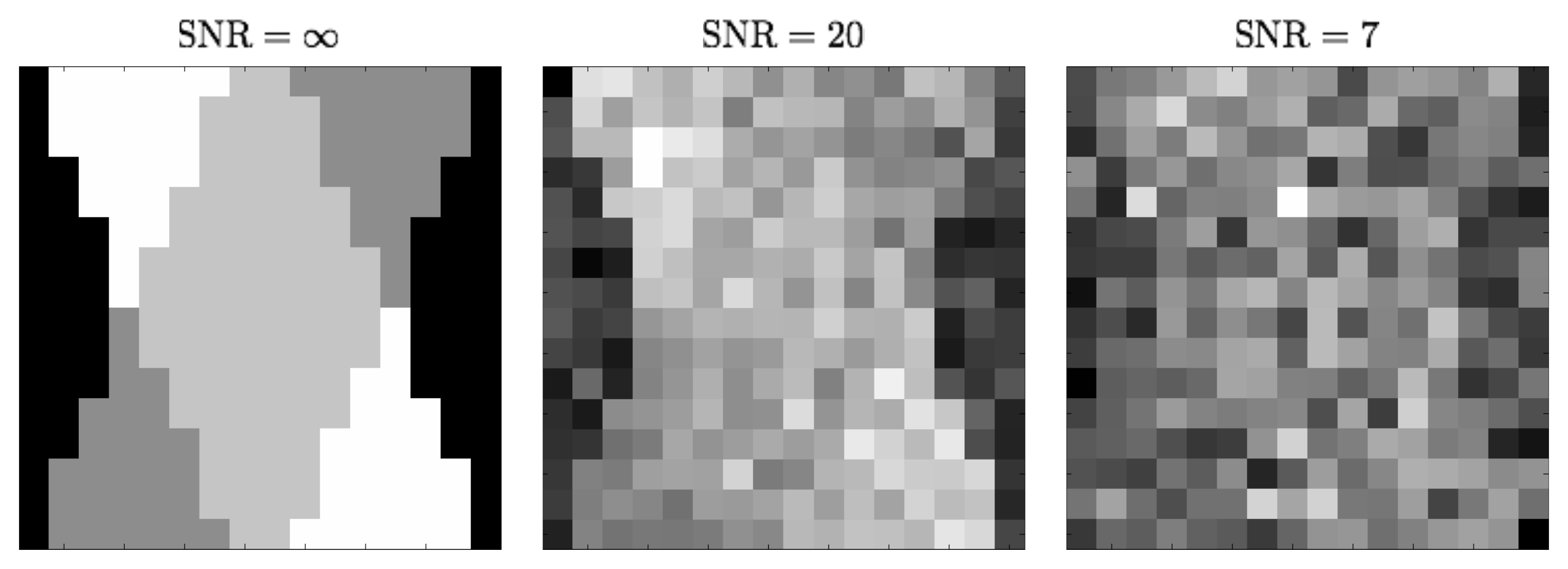} 
\caption{Example of a noise-free ``axial slice" of the simulated HARDI data (left) along with its noise contaminated versions for SNR = 20 (middle) and SNR =7 (right).}
\label{F1-2}
\end{figure}

The data simulation was repeated for two different values of $b$, {\it viz.} $b=1000$ s/mm$^2$ and $b=3000$ s/mm$^2$, with the corresponding SNRs of Rician noise\footnote{In the case at hand, SNR was defined to be a ratio of the {\em mean} amplitude of a noise-free HARDI signal to the standard deviation of the Gaussian noise contaminating the complex MR readout.} being 20 and 7. It should be noted that the above choice of $b$-values and their related SNRs is by no means arbitrary, but intended to imitate a real-life situation, in which increasing the value of $b$ comes at the price of a substantially reduced SNR. (The effect of Rician noise is exemplified in Fig.~\ref{F1-2} where a noise-free ``axial slice" of the simulated HARDI signal (left) is shown along with its noise contaminated versions for SNR = 20 (middle) and SNR=7 (right).) In such a case, the adverse effect of measurement noises could effectively counterpoise the gain in signal bandwidth which comes with exploiting higher $b$-values. In the context of SD, it is therefore important to understand the trade-off between the values of $b$ and SNR under various regularization schemes.    

Additionally, real-life HARDI data were acquired by means of a 3T Siemens TrioTim MRI scanner. The diffusion-encoded scans were collected from a 45 y.o. healthy volunteer at 2 mm isotropic resolution with: $b \in \{1000, 3000\}$ s/mm$^2$, TR=6300 ms, and TE = 85 ms. The transversal dimensions of the scans were equal to 128$\times$128 pixels, with a total of 50 axial slices used. The diffusion-encoding directions $\{\uu_k\}_{k=1}^K$ (with $K=64$) were defined using the method of generalized spirals \cite{Saff:1997ys}, which provides a closed-form solution to the problem of quasi-uniform sampling of $\Estu$.    

\subsection{Reference methods} 
In this work, we compared the performance of the proposed method against that of a number of alternative approaches. The first of these approaches was the positively constrained LS algorithm of \cite{Tournier:2008ve}, which will be referred below to as CSD. Note that this method disregards the presence of isotropic diffusion, and its solutions can be produced by solving \eqref{Cost} upon replacing $\Phi$ by $H$ and setting $\lambda = \mu = \nu = 0$. In the case when $\lambda > 0$, CSD transforms into the method of \cite{Jian:2007il}, which was the second reference method (referred below to as Min-$\mathcal{L}_1$) used in our comparative study. Note that, just like CSD, Min-$\mathcal{L}_1$ disregards {\em both} the presence of isotropic diffusion and the spatial-domain regularity of estimated fODFs. While still ignoring the effect of isotropic diffusion, the method of \cite{Reisert:2011ss} takes advantage of an original way to constrain the spatial behaviour of reconstructed fODFs using a fibre continuity model. This method (which we refer to below as CSD-FC) can be also described by the minimization problem in \eqref{Cost} upon replacing $\Phi$ by $H$ and setting $\lambda = \nu = 0$. Further, setting $\mu = 0$ in \eqref{Cost} would result in the SD algorithm recently proposed in \cite{Zhou:2014xy}. This method explicitly accounts for the presence of isotropic diffusion, while constraining the spatial regularity of its related IDM. However, the method of \cite{Zhou:2014xy} (referred below to as Min-TV-$\mathcal{L}_1$) does not enforce the spatial continuity of fibre orientations, as it is done in the case of the proposed algorithm. In \cite{Dellacqua:2010uq}, the authors proposed an SD approach based on the famed Lucy-Richardson deconvolution procedure (referred to as dRL). Although disregarding the spatial-domain regularity of the estimated fODFs, dRL had been particularly designed to suppress the influence of isotropic diffusion on the deconvolution results. Finally, for the sake of notational convenience, the SD method proposed in this paper will be referred below to as SCSD (standing for spatially constrained sparse deconvolution).  

\subsection{Comparison metrics}
To quantitatively assess the results produced by different SD routines, a total of four comparison metrics have been employed. The first of these metrics is {\em average angular error} (AAE) defined as
\begin{equation}\label{AAE}
{\rm AAE} = \frac{180^\circ}{\pi} \mathcal{E} \{ \arccos (\vv_0 \cdot \tilde{\vv}) \},
\end{equation}
where $\vv_0$ and $\tilde{\vv}$ stand for an original fibre orientation and its estimate respectively and $\mathcal{E}$ denotes the operator of statistical expectation (which is commonly approximated by a sample mean in practice). It should be noted that a standard way to infer the value(s) of $\tilde{\vv}$ from an estimated fODF is through finding the local maxima of the latter. Alternatively, one could use the estimated fODF to fit to it a model of the form $\sum_{l=1}^M \exp (\uu^T A_l \uu)$ (with $M$ denoting the number of fibres and $A_l$ being a symmetric $3\times 3$ matrix)  and then associate $\tilde{\vv}$ with the direction of the major eigenvector of $A_l$. When the number of fibres $M$ is known as {\it a priori} (as it is the case with computer-simulated studies), the second approach tends to produce more accurate results. For this reason, it was exploited in this paper.           
 
Additional metrics used in our comparative analysis were {\em true positive} (TP) and {\em false positive} (FP) rates of detection of the correct number of fibres at each voxel. Specifically, TP can be defined to be a percentage of data voxels, in which the number of local maxima of an estimated fODF coincided with the true number of simulated fibres. To render the computation of TP more robust towards overestimation errors, at each voxel the local maxima have been subjected to hard thresholding, set at the level of 20\% of their peak value. The FP metric, on the other hand, was particularly designed to assess the extent of overestimation of the true number of fibres \cite{Reisert:2011ss,Dellacqua:2010uq}. Formally, FP can be defined as
\begin{equation}\label{FP}
{\rm FP} = \mathcal{E} \{ (\tilde{M}-M_0)_+\},
\end{equation}
where $M_0$ and $\tilde{M}$ denote the true number of fibres and the number of local maxima in reconstructed fODFs. Note that, similarly to the case of TP, before evaluating FP, all the local maxima had been subjected to the same hard thresholding procedure.   

Finally, one of the principal outputs of the proposed method is an estimate of the IDM, which is a scalar-valued function of the spatial coordinate. In the case at hand, the quality of reconstructed IDMs can be assessed qualitatively based on the notion of image contrast. The latter can be defined as follows. First, we partition the entire image domain $\Omega$ into two subdomains, $\Omega_{in}$ and $\Omega_{out}$ (with $\Omega = \Omega_{in} \cup \Omega_{out}$), which encompass the regions occupied by the simulated fibres and purely isotropic diffusion (i.e., no fibres present), respectively. Also, we denote by $f_{iso}(\Omega_{in})$ and $f_{iso}(\Omega_{out})$ the restrictions of $f_{iso}$ to the corresponding subdomains. Then, letting $\mu_{in}$ and $\mu_{out}$ (resp. $\sigma_{in}$ and $\sigma_{out}$) denote the mean values (resp. the standard deviations) of $f_{iso}(\Omega_{in})$ and $f_{iso}(\Omega_{out})$, correspondingly, a measure of contrast $C$ can be defined as given by 
\begin{equation}\label{Contr}
C = 2 \, \frac{| \mu_{in} - \mu_{out} |}{\sigma_{in} + \sigma_{out}}.
\end{equation}
It is worthwhile noting that, the (real) IDMs used in the experimental study are piecewise constant ``images", in which case $\sigma_{in} = \sigma_{out} = 0$, resulting in $C = \infty$. Thus, in the case of estimated IDMs, it is reasonable to assume that higher values of $C$ represent more accurate estimates of the original IDMs.    

As a closing remark we note that, among all the SD methods under comparison, only Min-TV-$\mathcal{L}_1$ and SCSD are capable (by design) of computing the estimates of IDM, whereas neither of the remaining methods is endowed with the same capability. However, both CSD and Min-$\mathcal{L}_1$, as well as CSD-FC, have been proposed as means for estimation of the fODF $f_a$, in which case the remainder $H \tilde{f}_a - s$ (with $\tilde{f}_a$ being an estimate of $f_a$) is likely to represent estimation errors {\em along with} a contribution from $f_{iso}$. Furthermore, assuming the estimation errors to have a zero mean value, a reasonable estimate $\tilde{f}_{iso}$ of $f_{iso}$ can be obtained by averaging the remainder $H \tilde{f}_a - s$ over the spherical coordinate. Formally, one can define $\tilde{f}_{iso}$ to be
\begin{equation}\label{FISO}
\tilde{f}_{iso} \simeq \frac{A}{K} \, {\bf 1}^T \big( H \tilde{f}_a - s \big),
\end{equation}
where {\bf 1} denotes a $K$-dimensional (column) vector of ones and $A > 0$ is a proportionality constant that could be set to 1, since it does not have any effect on the value of $C$ in \eqref{Contr}. We also note that, by its design, the dRL algorithm of \cite{Dellacqua:2010uq} aims to recover an FOF (rather than an fODF) as given by \eqref{DECOMP}, thereby making the estimation of IDMs according to \eqref{FISO} inadequate. Consequently, quantitative comparisons in terms of $C$ have been performed for all the methods under consideration, except for dRL.     

\section{Results}\label{S6}
\subsection{Computer simulations}

\begin{figure}[!t]
\centering
\includegraphics[width=6in]{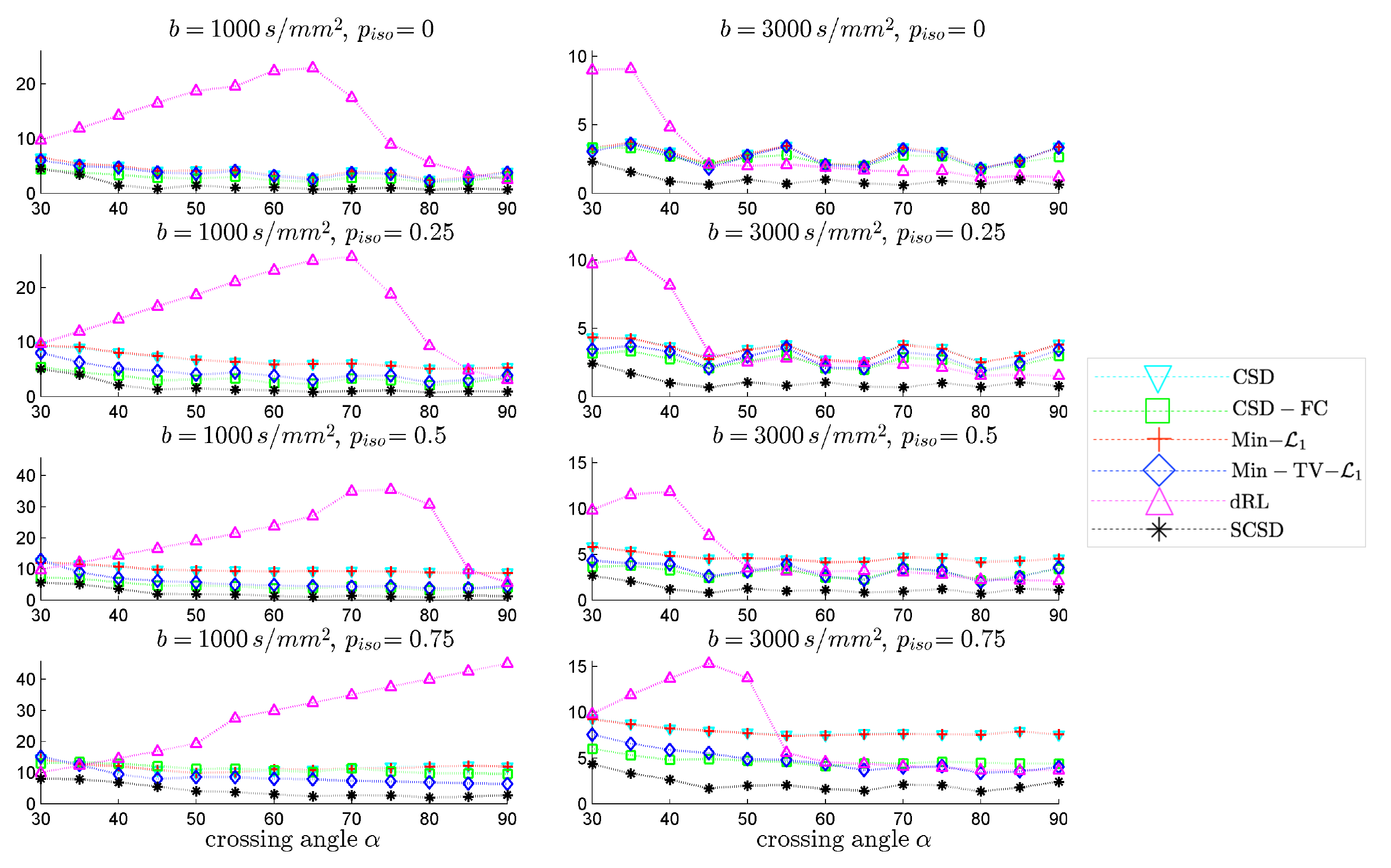} 
\caption{(Left columns of subplots) AAE produced by the tested methods for different $\alpha$, $p_{iso}$, and $b=1000$ s/mm$^2$; (Right columns of subplots) AAE produced by the tested methods for different $\alpha$, $p_{iso}$, and $b=3000$ s/mm$^2$.}
\label{F3}
\end{figure}

The left and right columns of subplots in Fig.~\ref{F3} show the values of AAE as a function of crossing angle $\alpha$, which have been obtained with the proposed and reference methods for $b=1000$ s/mm$^2$ and $b=3000$ s/mm$^2$, respectively. The regularization parameters for Min-$\mathcal{L}_1$, CSD-FC, Min-TV-$\mathcal{L}_1$, and SCSD were set to optimize the overall performance of the algorithms, namely: 1) $\lambda = 0.01$ for Min-$\mathcal{L}_1$, 2) $\mu = 0.01$ for CSD-FC, 3) $\lambda = 0.07$, $\nu = 0.01$ for Min-TV-$\mathcal{L}_1$, and 4) $\lambda = 0.03$, $\nu = 0.01$, $\mu = 0.4$ for SCSD. The dRL method has been reproduced following its description in \cite{Dellacqua:2010uq}. One can see that all the error curves, with the exception of that of dRL, exhibit the expected behaviour where AAE decreases with an increase in $\alpha$. Moreover, despite the substantially worse noise conditions for $b=3000$ s/mm$^2$, all the tested methods (again, with the exception of dRL) demonstrate better performance for $b = 3000$ s/mm$^2$, as compared to the case of $b=1000$ s/mm$^2$. In all the cases, however, the proposed SCSD method shows considerably better performance in comparison to the alternative solutions, with the ``second best" results produced by CSD-FC for lower values of $p_{iso}$ and by Min-TV-$\mathcal{L}_1$ for higher values of $p_{iso}$.    

\begin{figure}[!t]
\centering
\includegraphics[width=6in]{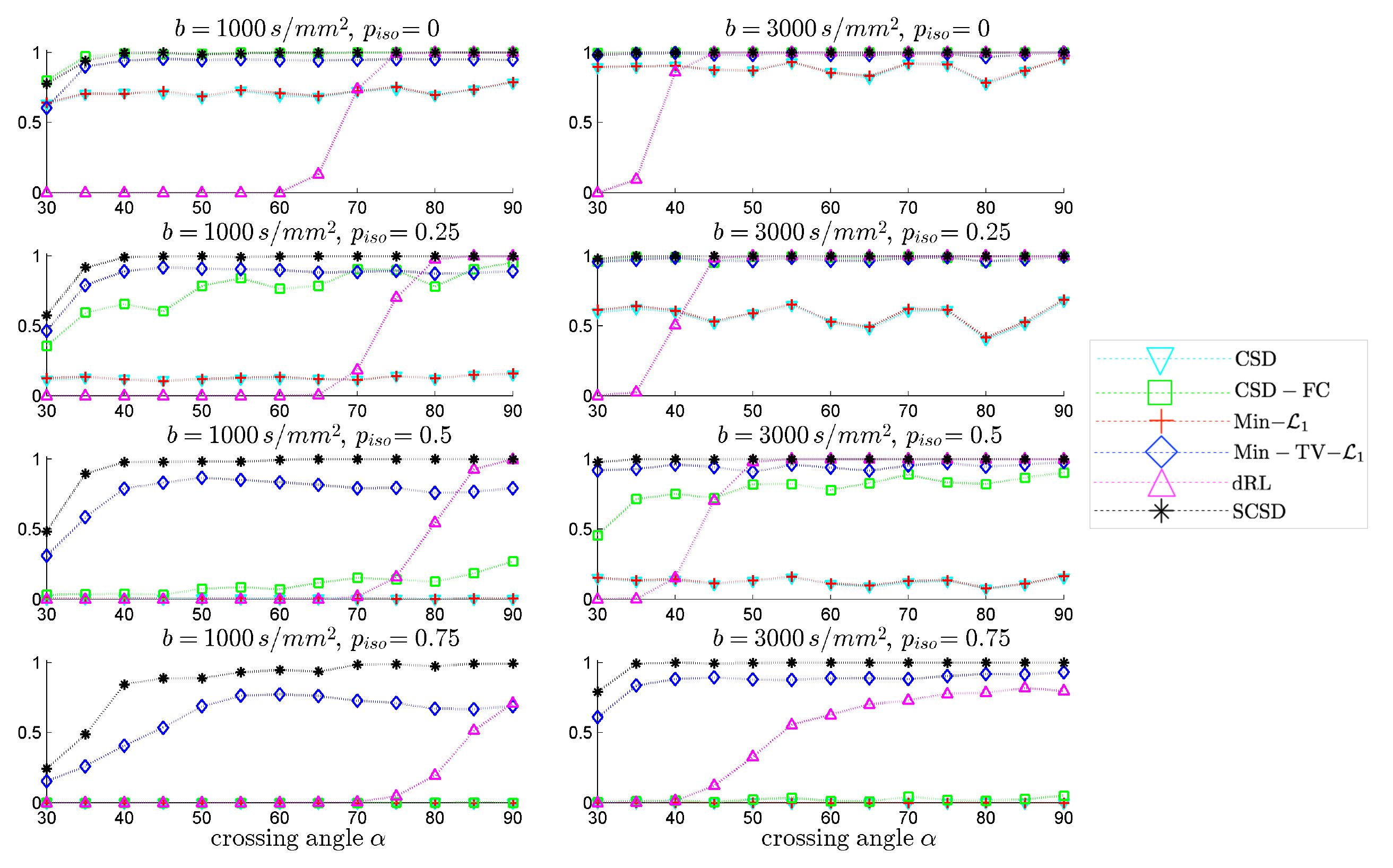} 
\caption{(Left columns of subplots) TP produced by the tested methods for different $\alpha$, $p_{iso}$, and $b=1000$ s/mm$^2$; (Right columns of subplots) TP produced by the tested methods for different $\alpha$, $p_{iso}$, and $b=3000$ s/mm$^2$.}
\label{F4}
\end{figure}

Before proceeding any further, one additional comment is in order regarding the behaviour of the AAE curves obtained with dRL. Specifically, one can see that, for $b=1000$ s/mm$^2$, the AAE is minimized for smaller values of $\alpha$, which is rather a counter-intuitive result. To understand why this happens, it is instructive to examine the behaviour of the TP curves produced by dRL (see Fig.~\ref{F4}). Specifically, one can see that for $b=1000$ s/mm$^2$ and $p_{iso} = 0$, dRL is incapable of resolving the crossing fibres of the numerical phantom for $\alpha \le 60^\circ$, with the resolvability problem becoming progressively worse with an increase in $p_{iso}$. In this case, the values of AAE effectively ``mirror" those of $\alpha$ up to the point when dRL starts detecting the correct number of fibres, after which AAE becomes a decreasing function of   the fibre crossing angle. It also deserves noting that for $b=3000$ s/mm$^2$, dRL demonstrates considerably improved performance in terms of AAE, even surpassing CSD and Min-$\mathcal{L}_1$ for $\alpha \ge 50^\circ$ and $p_{iso} \ge 0.5$. However, in all the alternative scenarios, the performance of dRL remains inferior to that of other methods under comparison. A possible explanation to this fact could be that the Lucy-Richardson algorithm exploited by dRL aims at recovering a maximum likelihood estimate under the assumption on measurement noise to be Poissonian. However, such a noise model can hardly be a good approximation to Rician distribution, which is inherent in MRI.

Fig.~\ref{F4} shows the values of TP which have been obtained using the proposed and reference methods for $b=1000$ s/mm$^2$  (left column of subplots) and $b=3000$ s/mm$^2$ (right column of subplots). One can see that, in the absence of isotropic diffusion (i.e., for $p_{iso} = 0)$, the proposed SCSD method performs comparably to CSD-FC for both values of $b$ (with Min-TV-$\mathcal{L}_1$ being the next ``best performer"). Yet, the moment $p_{iso}$ starts growing above zero, not only CSD and Min-$\mathcal{L}_1$, but also CSD-FC start loosing their ability to accurately estimate the true number of simulated fibres. Note that this is an expected result, considering the fact that neither of the above methods is endowed with facilities to explicitly account for the presence of isotropic diffision, in which case over-estimation errors become inevitable (see below). At the same time, for $b=3000$ s/mm$^2$, the proposed SCSD algorithm provides an ideal detection rate of one for $p_{iso} \in \{0, 0.25, 0.5\}$ and $\alpha \ge 30^\circ$, as well as for $p_{iso} = 0.75$ and $\alpha \ge 35^\circ$. It deserves noting that, owing to its properly accounting for the effect of isotropic diffusion, the performance of Min-TV-$\mathcal{L}_1$ is only marginally inferior to that of SCSD for $b=3000$ s/mm$^2$ and $p_{iso} \le 0.5$. One can also see that a nearly ideal TP rate is reached by the dRL algorithm as well. However, it does not happen until after $\alpha$ reaches relatively large values (e.g., for $\alpha \ge 50^\circ$ with $p_{iso} = 0.5$ and $b=3000$ s/mm$^2$).

\begin{figure}[!t]
\centering
\includegraphics[width=6in]{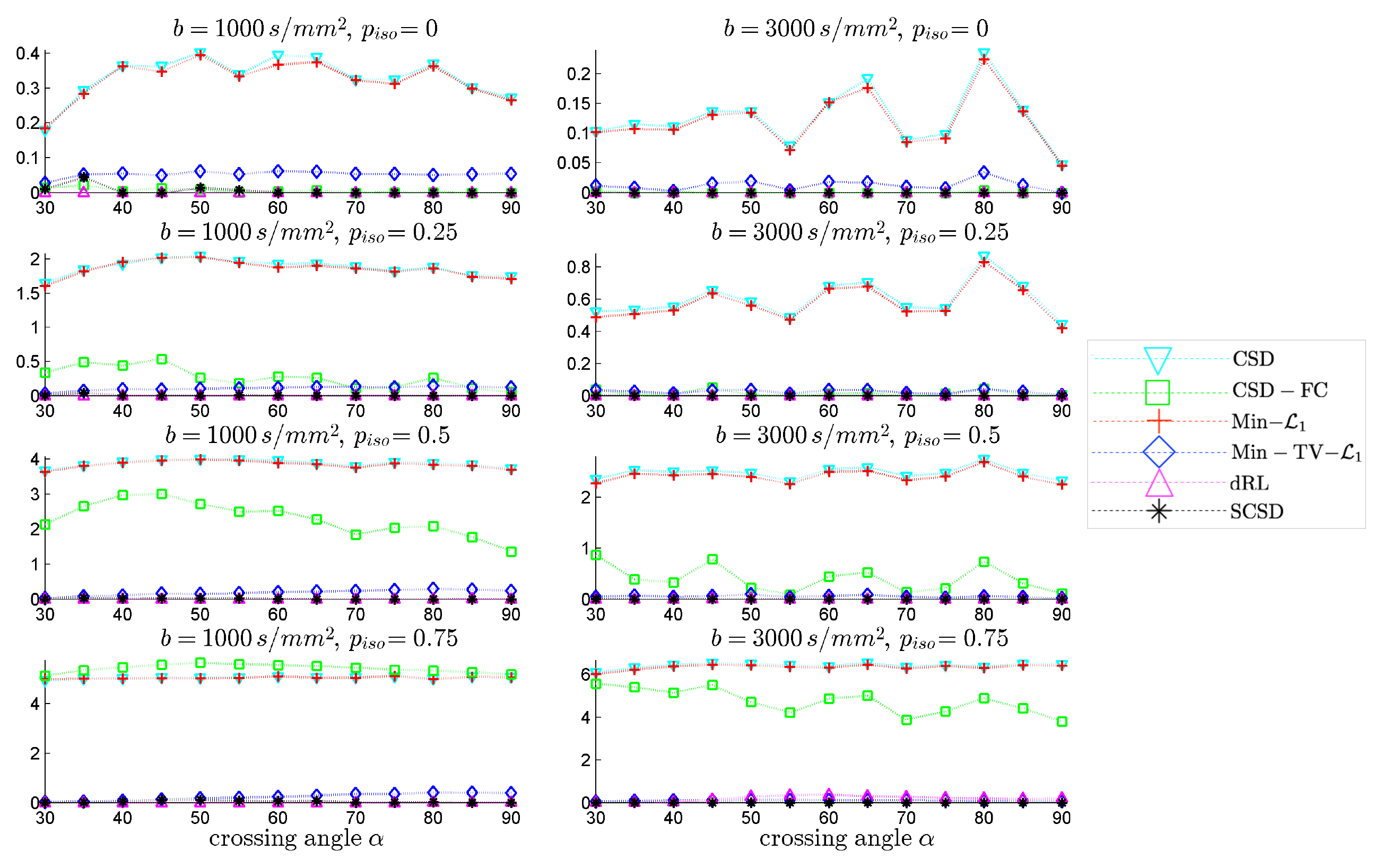} 
\caption{(Left columns of subplots) FP produced by the tested methods for different $\alpha$, $p_{iso}$, and $b=1000$ s/mm$^2$; (Right columns of subplots) FP produced by the tested methods for different $\alpha$, $p_{iso}$, and $b=3000$ s/mm$^2$.}
\label{F5}
\end{figure}

One of the principle applications of HARDI is in multi-fibre tractography, in which case SD can be used to improve the resolvability of multiple fibre tracts within each given voxel in a region of interest (ROI). In this situation, overestimating the number of fibres is likely to produce spurious fibre tracts, thereby rendering the resulting reconstructions unreliable. Unfortunately, unless properly regularized, some SD routines tend to amplify the effect of noise, which in turn results in numerous false (local) maxima in reconstructed fODFs -- the maxima that can be easily confused with the true modes of the latter. For this reason, it is important to compare the performance of SD routines in terms of the FP metric. Such comparative results are summarized in Fig.~\ref{F5}, the left and right columns of which correspond to the cases of $b=1000$ s/mm$^2$ and $b=3000$ s/mm$^2$, respectively. One can see that, in this case, the worst results are produced by CSD and Min-$\mathcal{L}_1$ for all values of $b$ and $\alpha$. The CSD-FC algorithm, on the other hand, provides a close to zero FP rate for $b=3000$ s/mm$^2$ and $p_{iso} \in \{0, 0.25\}$. Unfortunately, its performance deteriorates for higher values of $p_{iso}$ (which is particularly noticeable for $b=1000$ s/mm$^2$). Surprisingly, dRL seems to provide an ideal FP rate of zero for all values of $b$ and $\alpha$ under consideration. However, from our analysis of Fig.~\ref{F4} it is not hard to see that it happens only because this method tends to underestimate the true number of simulated fibres. Finally, one can also see that, for $b=3000$ s/mm$^2$, Min-TV-$\mathcal{L}_1$ provides a fairly small FP rate, whereas the proposed SCSD algorithm succeeds to attain the ideal FP rate of zero for all values of $p_{iso}$ and $\alpha$.

\begin{figure}[!t]
\centering
\includegraphics[width=6in]{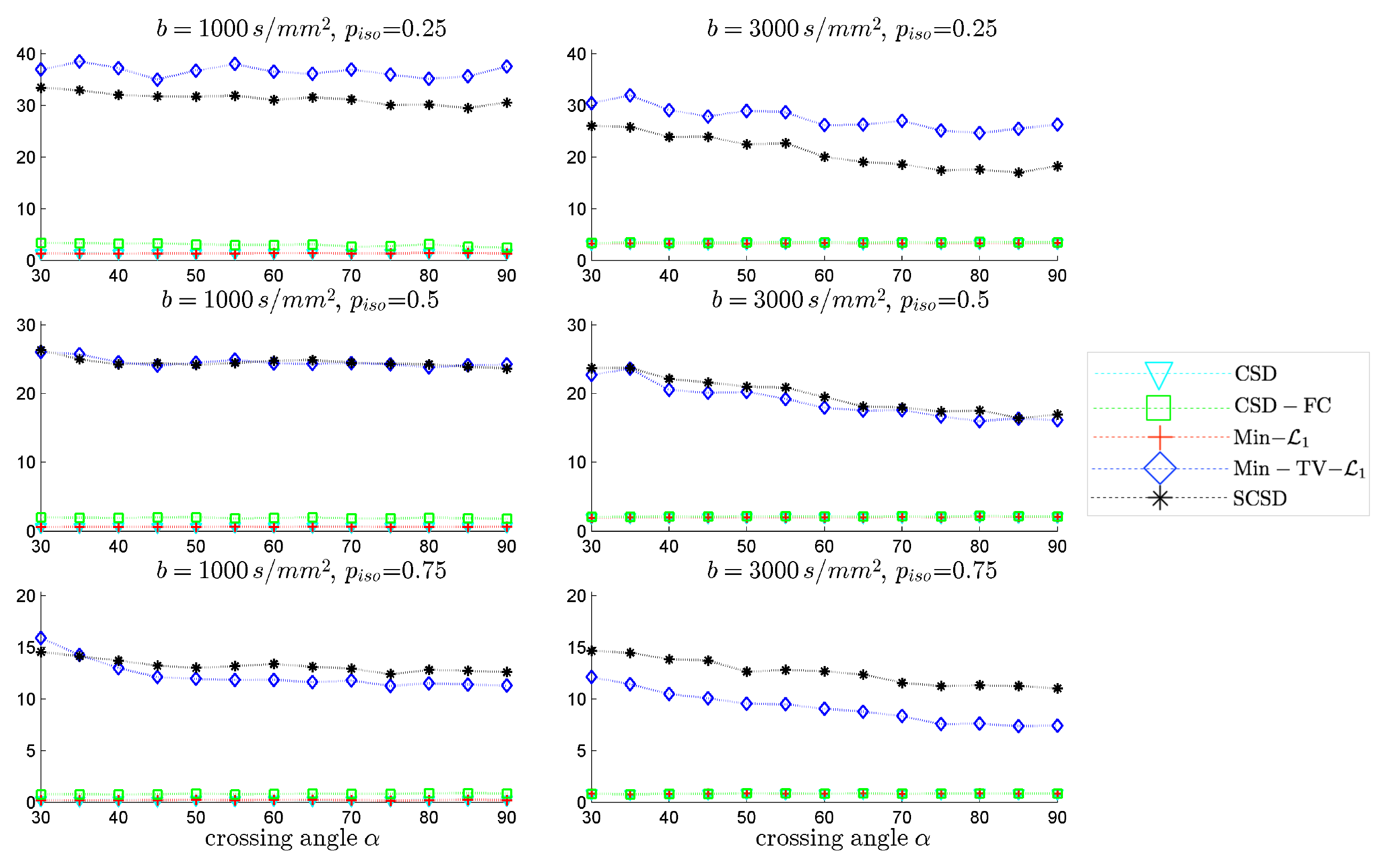} 
\caption{(Left columns of subplots) C produced by the tested methods for different $\alpha$, $p_{iso}$, and $b=1000$ s/mm$^2$; (Right columns of subplots) C produced by the tested methods for different $\alpha$, $p_{iso}$, and $b=3000$ s/mm$^2$.}
\label{F6}
\end{figure}

\begin{figure}[!t]
\centering
\includegraphics[width=5in]{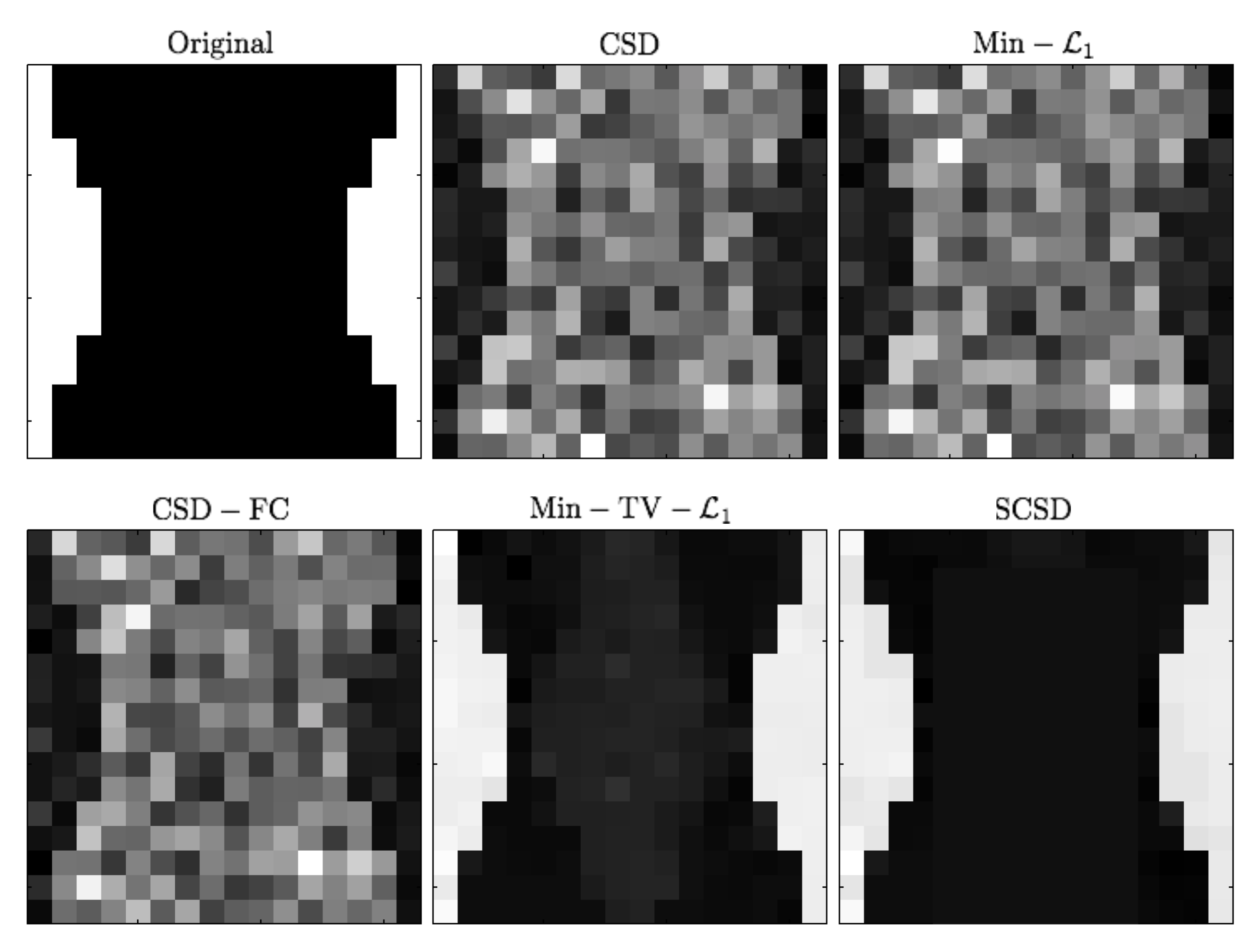} 
\caption{An ``axial" slice of the IDMs recovered by different SD methods under comparison for $b=3000$ s/mm$^2$, $\alpha = 45^\circ$, and $p_{iso}=0.5.$}
\label{F7}
\end{figure}

The results of our final quantitative comparison are summarized in Fig.~\ref{F6}, which shows the values of contrast $C$ obtained using different SD methods under comparison for $b=1000$ s/mm$^2$ (left column of subplots) and $b=3000$ s/mm$^2$ (right column of subplots). Predictably enough, the best contrast is achieved by the Min-TV-$\mathcal{L}_1$ and SCSD algorithms, owing to their inherent ability to account for the presence of isotropic diffusion. Moreover, out of the two, the proposed SCSD algorithm yields the higher values of $C$ for all simulated scenarios. An additional illustration of the effect of incorporation and spatial  regularization of the isotropic diffusion component is provided in  Fig.~\ref{F7}, which depicts a 2-D ``axial" slice of the IDMs reconstructed by different SD methods under comparison for $b=3000$ s/mm$^2$, $\alpha = 45^\circ$, and $p_{iso} = 0.5$. (Note that, for the sake of the clarity of visualization, the IDMs in Fig.~\ref{F7} have been normalized so as to make their minimum and maximum values correspond to black and white pixel values, respectively.) One can see that the IDM reconstruction produced by SCSD is virtually indistinguishable from the original IDM (as shown in the upper, leftmost subplot of the figure), with the second best result produced by the Min-TV-$\mathcal{L}_1$ algorithm. At the same time, neither CSD, Min-$\mathcal{L}_1$ nor CSD-FC can attain a comparable accuracy of estimation of the spatial pattern of isotropic diffusion, as represented by the original IDM. In particular, even though their respective reconstructions do bear some global resemblance to the original IDM, the level of estimation errors is too high to deem these reconstructions useful.

\subsection{In vivo data experiments}

\begin{figure}[!t]
\centering
\includegraphics[width=6in]{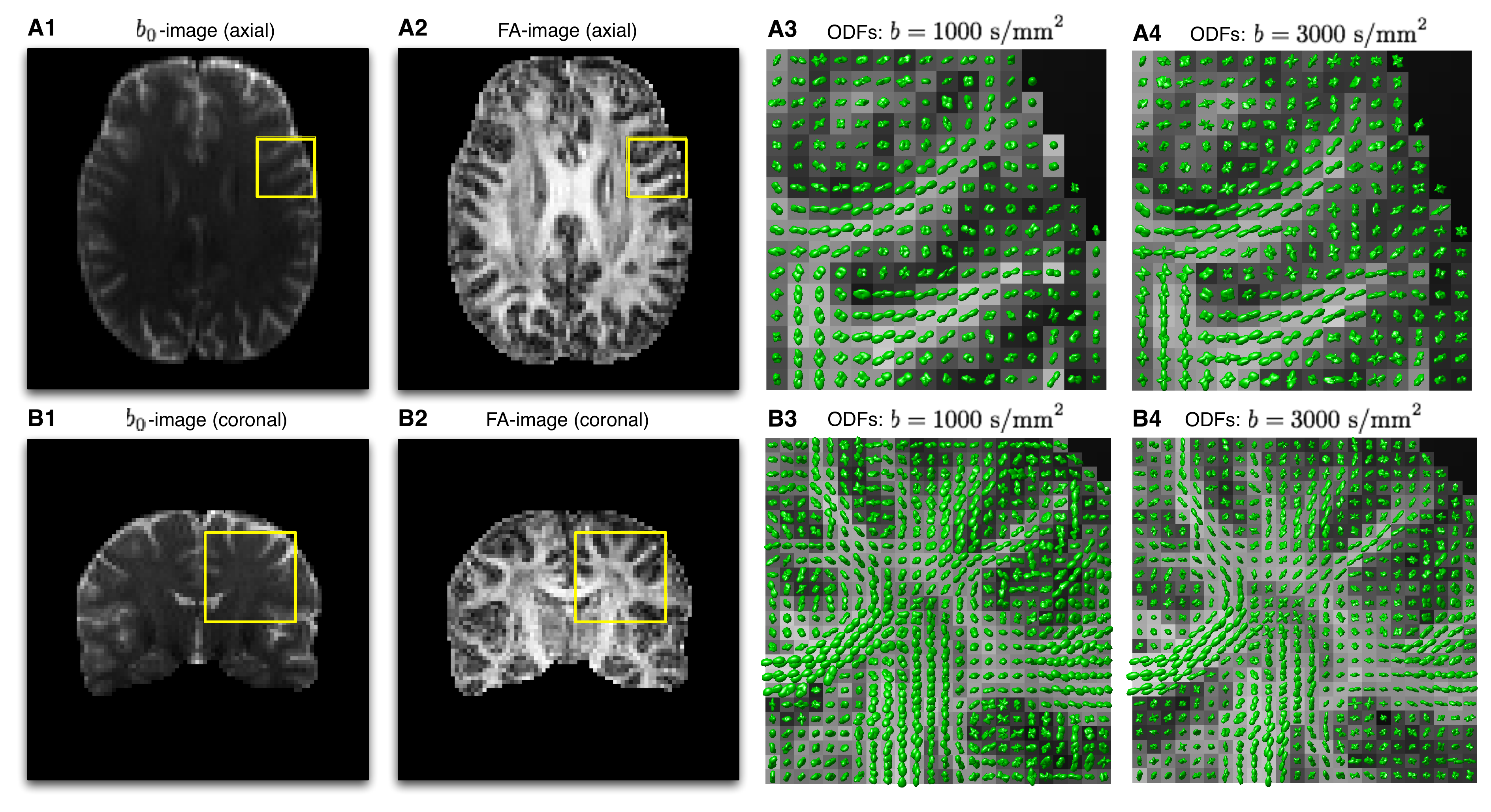} 
\caption{(Subplots A1-A2) $b_0$-image and FA images of an axial slice of an {\it in vivo} scan used for presentation of experimental results; (Subplots A3-A4) ODFs corresponding to the localized region indicated by the yellow squares in Subplots A1-A2 for $b=1000$ s/mm$^2$ and $b=3000$ s/mm$^2$, respectively;  (Subplots B1-B2) $b_0$-image and FA images of a coronal slice from the same scan; (Subplots B3-B4) ODFs corresponding to the localized region indicated by the yellow squares in Subplots B1-B2 for $b=1000$ s/mm$^2$ and $b=3000$ s/mm$^2$, respectively.}
\label{F8}
\end{figure}

As the next step of our experimental study, real-life estimation has been performed using {\it in vivo} diffusion data (see Section~\ref{S5-1}). For the sake of the clarity of visualization, we restrict demonstration of the obtained reconstructions to 2-D views (aka ``slices"), two examples of which are depicted in Fig~\ref{F8}. In particular, Subplots A1-A2 of the figure display an axial slice of the acquired $b_0$-volume and its corresponding fractional anisotropy (FA) image, respectively, while Subplots B1-B2 show a coronal slice from the same 3-D volume along with its associated FA image, in the same order. Additionally, Subplots A3-A4 of Fig.~\ref{F8} show the ODFs corresponding to the localized regions indicated by the yellow squares in Subplots A1-A2 for the case of $b=1000$ s/mm$^2$ and $b=3000$ s/mm$^2$, respectively. (To facilitate the analysis, the ODFs are shown over the background of their associated FA values.) Analogous results pertaining to the coronal view are shown in Subplots B3-B4 of the figure. It should be noted that, in both cases, the ODFs have been computed by means of the FRACT algorithm of \cite{Haldar:2013fk}, which seems to provide a reasonable balance between the robustness of more traditional QBI \cite{Descoteaux:2007jw} and the high resolution gain of its solid-angle formulation \cite{Aganj:2010fk}.   

\begin{figure}[!t]
\centering
\includegraphics[width=6in]{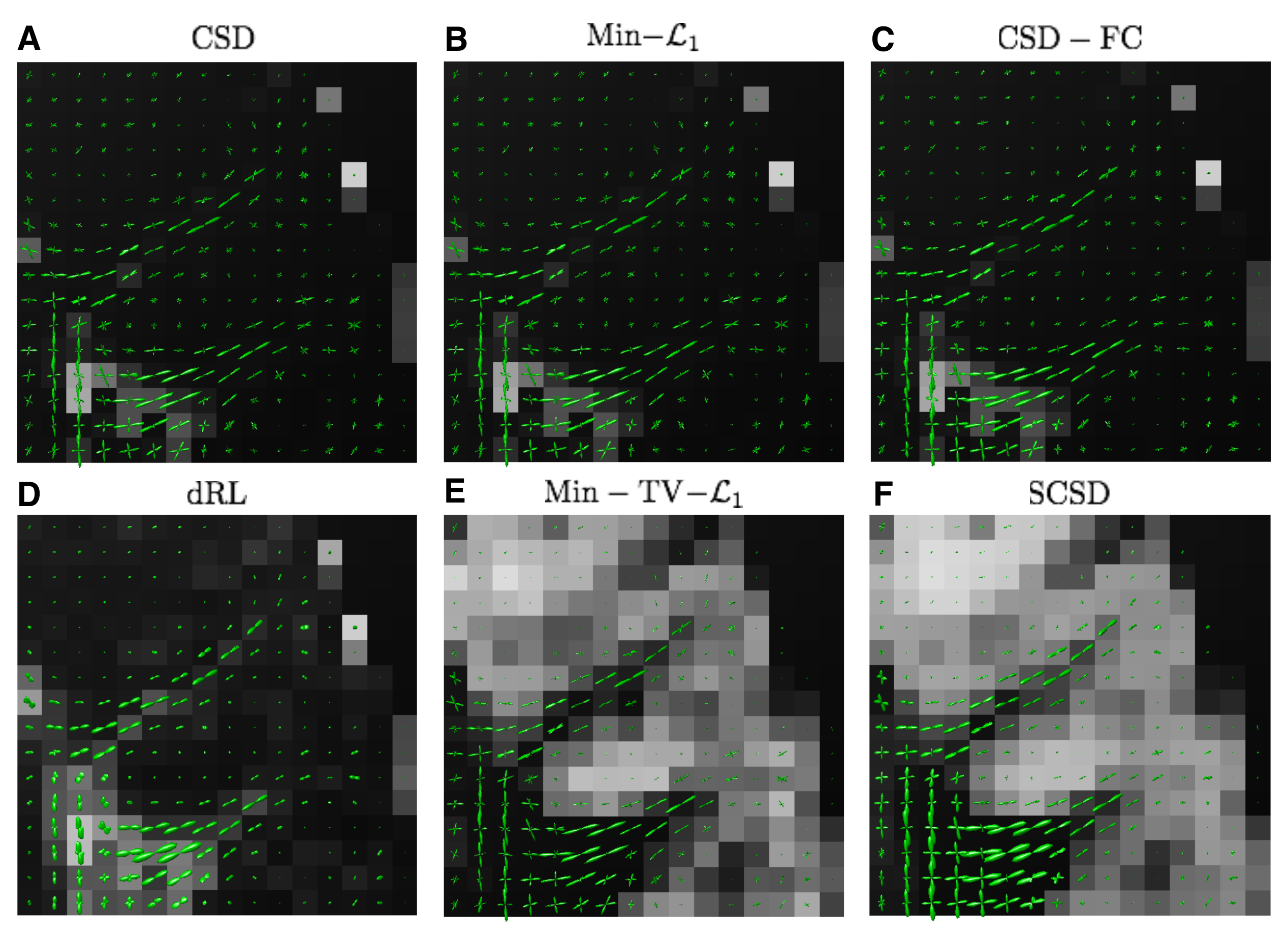} 
\caption{Estimated fODFs computed by CSD (A), Min-$\mathcal{L}_1$ (B), CSD-FC (C), dRL (D), Min-TV-$\mathcal{L}_1$ (E), and SCSD (F) for the axial view and $b=1000$ s/mm$^2$. The fODFs are superimposed over the background of their related IDMs.}
\label{F9}
\end{figure}

\begin{figure}[!t]
\centering
\includegraphics[width=6in]{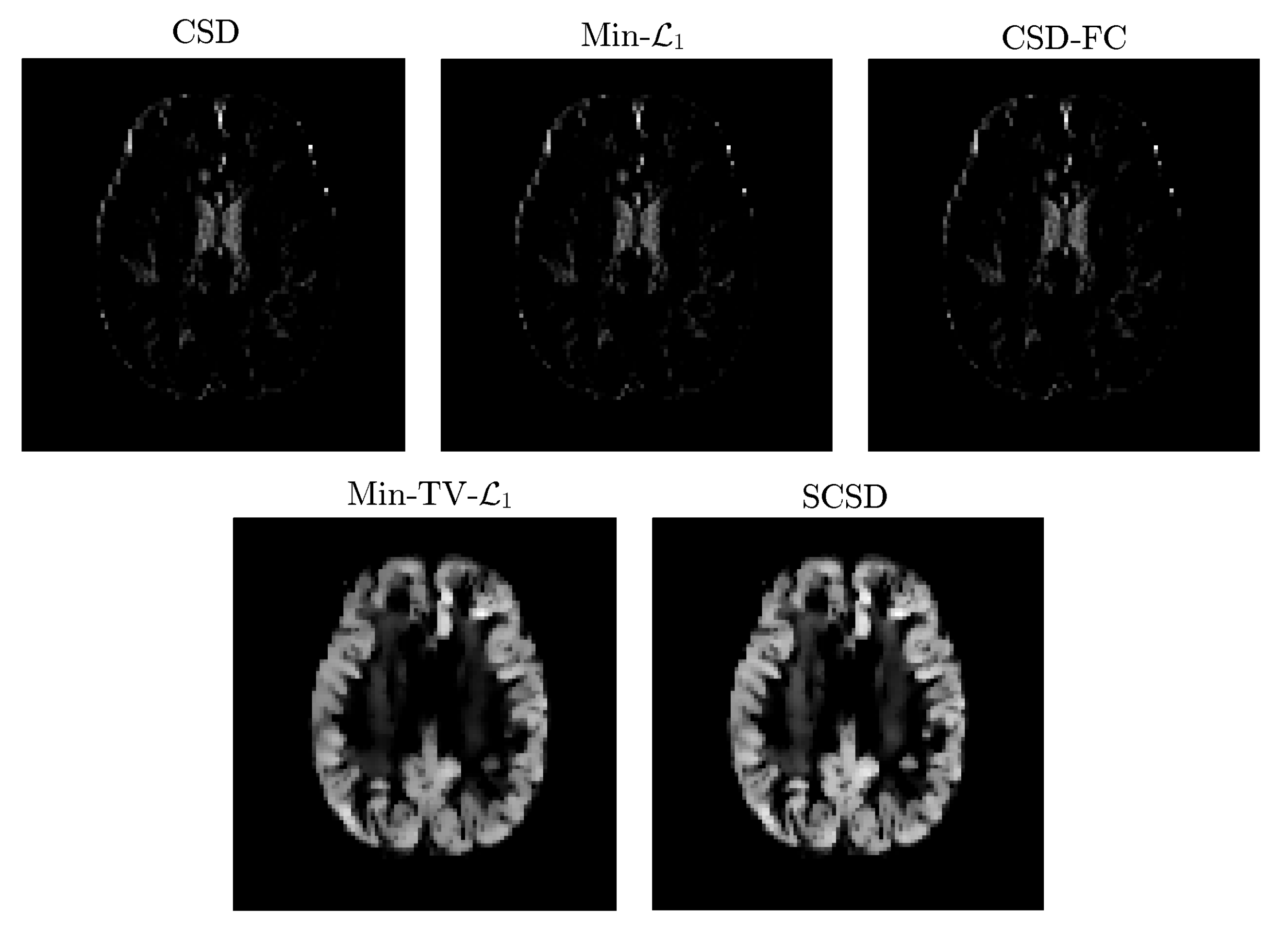} 
\caption{``Axial" IDMs estimated by various SD methods for $b=1000$ s/mm$^2$.}
\label{F13}
\end{figure}

Fig.~\ref{F9} depicts the reconstructions of fODFs obtained using the CSD (Subplot A), Min-$\mathcal{L}_1$ (Subplot B), CSD-FC (Subplot C), dRL (Subplot D), Min-TV-$\mathcal{L}_1$ (Subplot E), and SCSD (Subplot F) for the axial view and $b=1000$ s/mm$^2$. As opposed to Subplots A3-4 and B3-4 in Fig.~\ref{F8}, the fODFs in Fig.~\ref{F9} are shown superimposed over the values of their corresponding IDMs. Analysing these results reveals the principal drawback of SD methods which disregard the effect of isotropic diffusion. In particular, not only CSD and Min-$\mathcal{L}_1$ but also CSD-FC tend to yield spurious estimates of fODFs in anatomical regions corresponding to the cortical grey matter -- the result which stands at odds with the fact that these regions are known to be devoid of neural fibre bundles. Further, although being capable of coping with the presence of isotropic diffusion, the dRL algorithm is ``blending" $f_a$ and $f_{iso}$, which effectively impairs the angular resolution, and therefore the resolvability of crossing fibre tracts. Moreover, neither of the aforementioned methods has been found to be capable of reliably recovering the IDMs. At the same time, both Min-TV-$\mathcal{L}_1$ and SCSD yield anatomically consistent reconstructions of the IDMs, with much less noisy results obtained in the case of SCSD. (This point is further illustrated by Fig.~\ref{F13} which shows the ``zoomed-out" IDMs recovered by the SD methods under comparison for the case of $b=1000$ s/mm$^2$.) Moreover, a closer inspection of Fig.~\ref{F9} reveals that the fODFs estimated by means of SCSD are characterized by a smoother and more consistent spatial variability (owing to the fibre continuity constraint), thereby exhibiting a better adherence to the expected connectivity within an {\it in vivo} brain.

\begin{figure}[!t]
\centering
\includegraphics[width=6in]{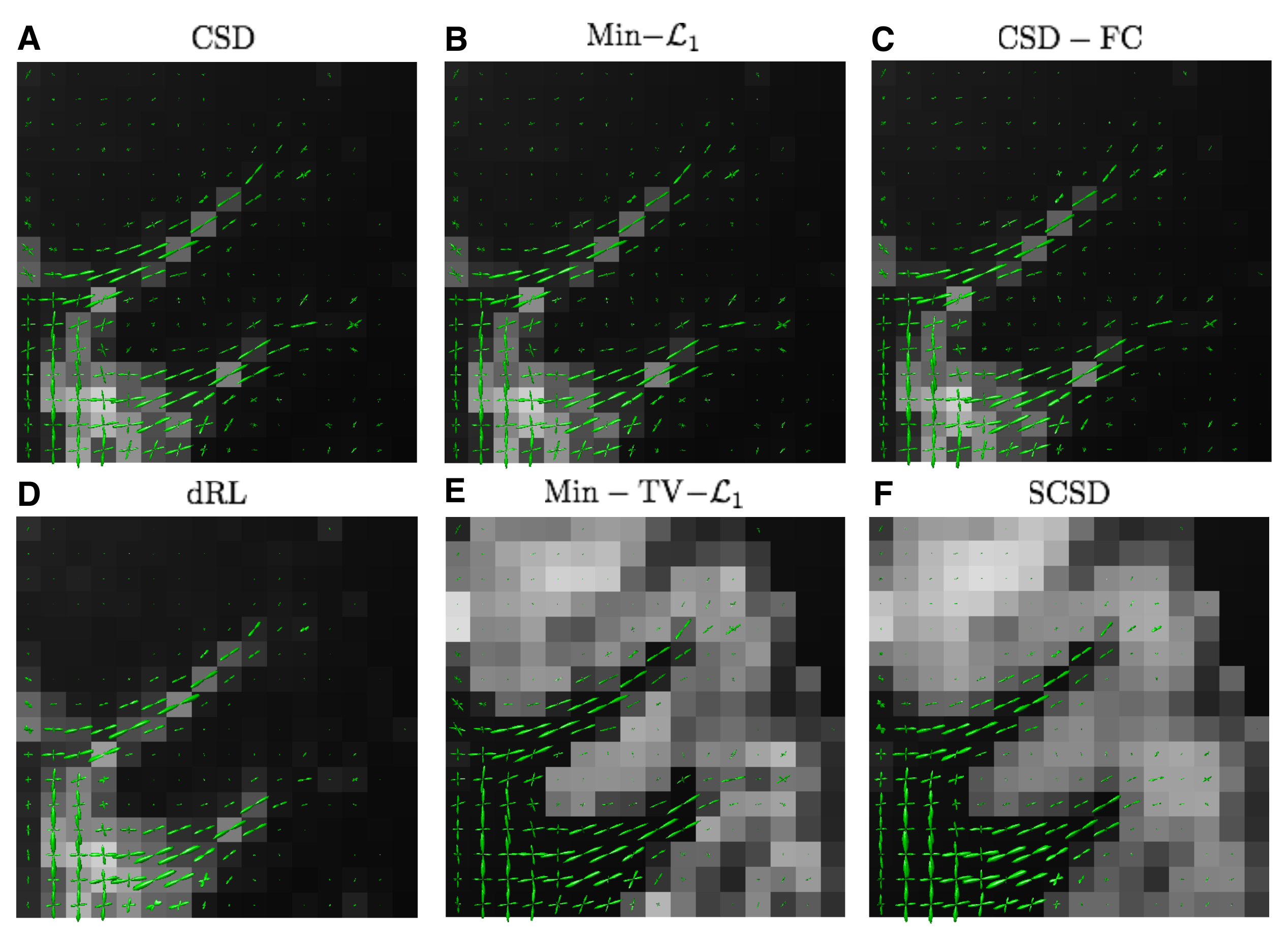} 
\caption{Estimated fODFs computed by CSD (A), Min-$\mathcal{L}_1$ (B), CSD-FC (C), dRL (D), Min-TV-$\mathcal{L}_1$ (E), and SCSD (F) for the axial view and $b=3000$ s/mm$^2$. The fODFs are superimposed over the background of their related IDMs.}
\label{F10}
\end{figure}

\begin{figure}[!t]
\centering
\includegraphics[width=6in]{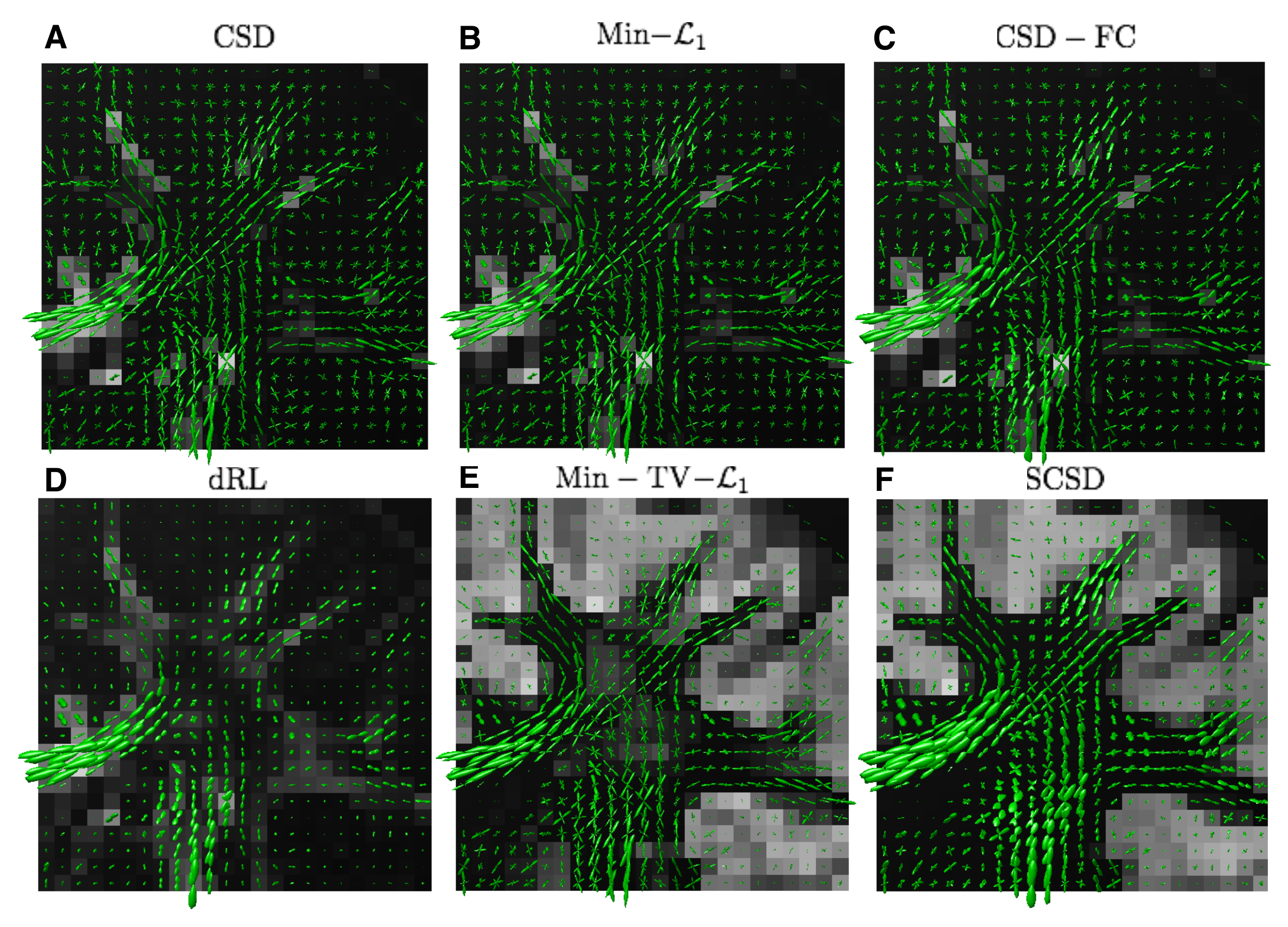} 
\caption{Estimated fODFs computed by CSD (A), Min-$\mathcal{L}_1$ (B), CSD-FC (C), dRL (D), Min-TV-$\mathcal{L}_1$ (E), and SCSD (F) for the coronal view and $b=1000$ s/mm$^2$. The fODFs are superimposed over the background of their related IDMs.}
\label{F11}
\end{figure}

\begin{figure}[!t]
\centering
\includegraphics[width=6in]{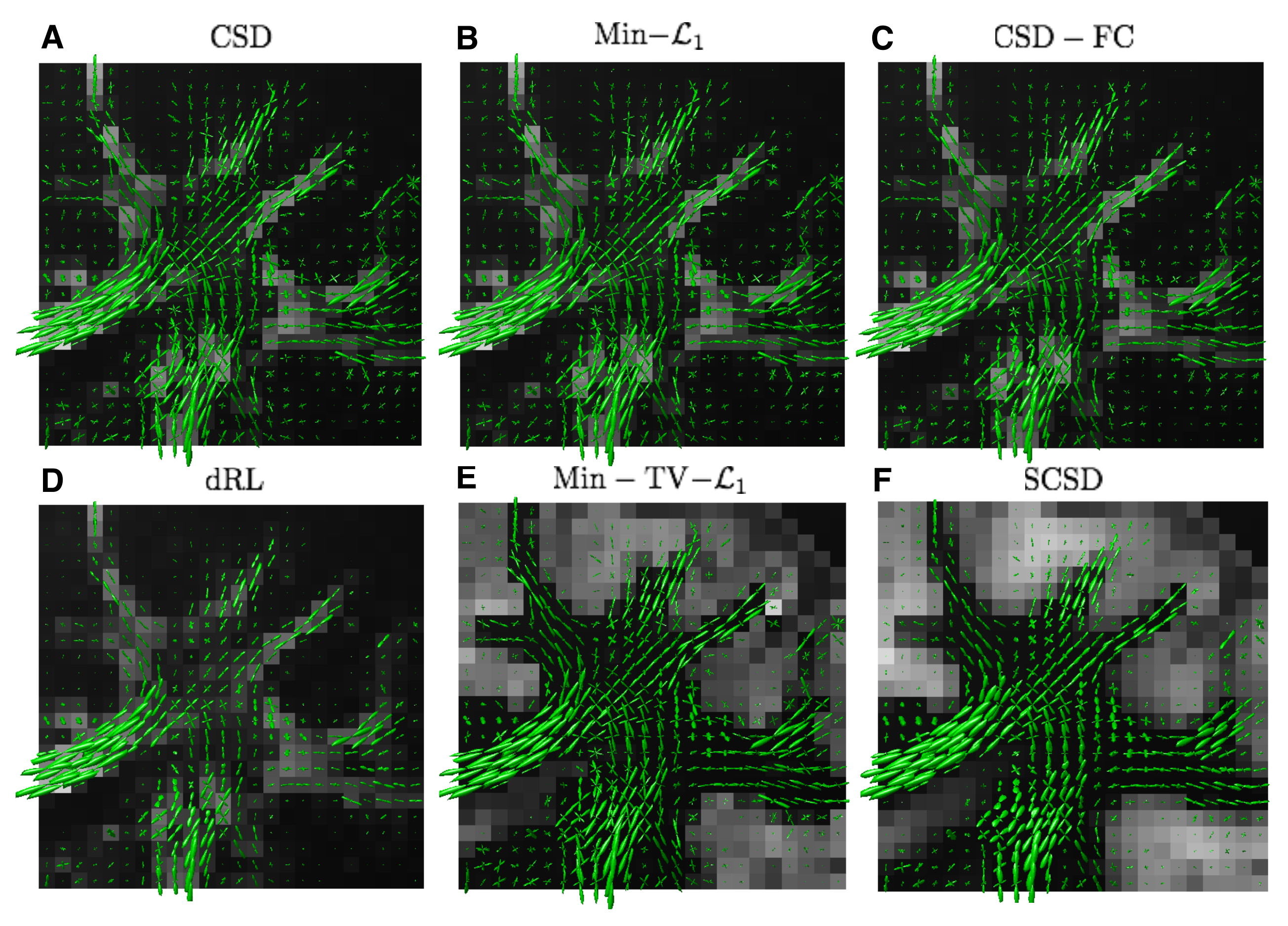} 
\caption{Estimated fODFs computed by CSD (A), Min-$\mathcal{L}_1$ (B), CSD-FC (C), dRL (D), Min-TV-$\mathcal{L}_1$ (E), and SCSD (F) for the coronal view and $b=3000$ s/mm$^2$. The fODFs are superimposed over the background of their related IDMs.}
\label{F12}
\end{figure}

The reconstructions obtained for the same axial view and $b=3000$ s/mm$^2$ are depicted in Fig.~\ref{F10}, whose composition is identical to that of Fig.~\ref{F9}. Although fairly close in appearance to the previous case, these reconstructions allow us to make a number of important observations. First of all, as compared to the case of $b=1000$ s/mm$^2$, a wider bandwidth of the HARDI signals at $b=3000$ s/mm$^2$ leads to a better angular resolution, which is particularly noticeable in the case of dRL. Moreover, despite considerably worse noise conditions, the fODF reconstructions obtained by means of Min-TV-$\mathcal{L}_1$ and SCSD have much less residual noise over the areas occupied by cortical grey matter, where isotropic diffusion is expected to prevail. This fact indicates the effectiveness of the regularization schemes exploited by these SD methods. Finally, a closer inspection of the glyphs in Fig.~\ref{F10} reveals that the fODF reconstructions yielded by SCSD demonstrate a better spatial smoothness and anatomical consistency. The same observations can be made in the case of the coronal view, as shown in Figs. \ref{F11} and \ref{F12} for $b=1000$ s/mm$^2$ and $b=3000$ s/mm$^2$, respectively. (Note that the local region represented by these figures has been chosen according to the results in \cite{Dellacqua:2010uq}, which also provides indication of specific fibre bundles within the selected regions of interest). 

\section{Discussion and conclusions}\label{S7}
In the present paper, we introduced a novel approach to the problem of non-blind SD of HARDI data. As opposed to many alternative methods of SD, the proposed algorithm can explicitly account for the effect of isotropic diffusion, which makes it capable of performing reliably across the entire brain, thereby avoiding the need to restrict the computations to the regions of white matter. In particular, in addition to reconstruction of fODFs $f_a(\uu|\rr)$, the algorithm can also yield a useful estimation of its related IDM $f_{iso}(\rr)$, which quantifies a relative contribution of the isotropic diffusion component as well as its spatial pattern. Moreover, one of the principal contributions of the present work is to demonstrate the effectiveness of exploiting {\em different} prior models for regularization of the spatial-domain behaviour of the reconstructed fODFs and IDMs. Specifically, the fibre continuity model of \cite{Reisert:2011ss} has been used to force the local maxima of the fODFs to vary consistently throughout the brain, whereas the bounded variation model of \cite{Rudin:1992fh} has helped us to achieve piecewise smooth reconstructions of the IDMs (which appear to be in a good agreement with brain anatomy). Moreover, although the results of our experimental study support the above prior models, the latter are by no means exclusive, and hence additional improvements could still be achieved through the use of more advanced methods of spatial regularization. Thus, for example, the bounded variation model could have been replaced by requiring the diffusion-encoded images $s_k(\rr)$ to have sparse representation coefficients in the domain of a multi-resolution transform (e.g., framelets \cite{Daubechies:2003xx}). At the same time, the linear filtering resulting from the fibre continuity assumption (as explained in the Appendix) could have been potentially replaced by a non-linear spatially-adaptive filtering. Whether or not the above modifications can result in substantial improvements in the quality of SD reconstruction, it remains a subject of our further research. 

The estimated IDMs could serve as an additional input to fibre tractography or an independent observation for further statistical inference. In both scenarios, we believe the spatial regularization is essential to make the reconstructions of fODFs and IDMs be in a better agreement with the anatomical structure of the brain. Furthermore, as long as fibre tractography is concerned, our experimental results indicate the possibility to achieve a better angular resolution using higher values of $b$ (e.g., $b=3000$ s/mm$^2$). It is important to note that this outcome would not have been possible without regularizing the IDMs (as explained in Section~\ref{S4} and experimentally verified in Section~\ref{S6}) due to the adverse effect of measurement noises.  

The proposed SCSD algorithm has been formulated as a convex minimization problem, which admits a unique and stable minimizer. Moreover, an important contribution of this work is to introduce a computationally efficient scheme for numerical implementation of SCSD. In particular, using ADMM, we have been able to find the optimal solution via a sequence of simpler optimization problems, which are both computationally efficient and amenable to parallel computations. 

The performance of the SCSD method has been quantitatively compared against that of a number of alternative approaches, which have been specially selected to demonstrate the importance of various assumptions and constraints. Although SCSD has been shown to outperform the reference methods in terms of all the comparison metrics used, it still has a few limitations, whose mitigation could potentially produce more accurate reconstructions. Thus, from a Bayesian perspective, the cost functional in \eqref{Cost} suggests the measurement noise to be of an additive Gaussian nature, while in practice it is more likely to be Rician. It should be noted, however, that the Gaussian model can serve as a good approximation to the Rician model for relative high values of SNR (i.e., SNR $\gg 5$) \cite{Dolui:2012fk}. Besides, using the Gaussian model is essential for rendering the optimization problem convex and computationally tractable. Yet, we note that, if using the Rician statistics is preferred, it should be straightforward to modify the SCSD algorithm as detailed in \cite{Dolui:2012fk} for a similar setting. Finally, it should also be mentioned that the present version of SCSD assumes the diffusion outside white matter to be purely isotropic, while evidence exists that the diffusion within grey matter could exhibit a certain degree of anisotropy \cite{Rathi:2013ab}. To account for this phenomenon, one could, for instance, replace the constant column of $\Phi$ in \eqref{Pmatrix} by spherical harmonics up to second degree inclusive. Exploring such a modification and its effect on the quality of SD reconstruction defines another direction of our future research.

\section*{Appendix}
To find a computationally efficient way to solve \eqref{VJ}, it is convenient to slightly simplify the notations first. To this end, we note that the problem has a general form of 
\begin{equation}\label{L2AP}
\min_w  \left\{ \frac{1}{2}\| w - q \|_2^2 + \tau \| {\bf T}_\vv w \|_2^2 \right\},
\end{equation}
with $w$ interpreted as an $I$-dimensional (row) vector to be optimized over, $q$ is a data vector, $\tau > 0$ is a fixed regularization constant, and ${\bf T}_\vv: \mathbb{R}^I \to \mathbb{R}^I$ is the operator of {\em directional differencing} in the direction of $\vv \in \Estu$. The problem \eqref{L2AP} is a standard LS problem, whose optimal solution $w^\star$ satisfies a system of normal equations of the form
\begin{equation}\label{L2SOL}
\left( {\bf I} + 2 \tau \, {\bf T}_\vv^\ast {\bf T}_\vv \right) w^\star = q,
\end{equation}
with ${\bf T}_\vv^\ast$ standing for the adjoint of ${\bf T}_\vv$. 

It goes without saying that a straightforward inversion of the matrix on the left-hand side of \eqref{L2SOL} cannot be accepted as a practical option, even for moderately sized diffusion-encoded images. Thus, a more efficient way to solve \eqref{L2SOL} needs to be found. To this end, we first note that, if the practical computation of partial differences was based on a standard backward-differencing scheme, then multiplication by ${\bf T}_\vv$ in the spatial domain would be equivalent to linear filtering with frequency response $H_\vv(\boldsymbol\omega)$ given by
\begin{equation}\label{IFILT1}
H_\vv(\boldsymbol\omega) = 2 \sum_{d=1}^3 \vv_d \sin \frac{\boldsymbol\omega_d}{2} e^{-\jmath \left( \frac{\boldsymbol\omega_d - \pi}{2} \right)},  \quad \boldsymbol\omega \in \mathbb{R}^3,
\end{equation}
with $\vv = (\vv_1, \vv_2, \vv_3)$ and $\boldsymbol\omega = (\boldsymbol\omega_1, \boldsymbol\omega_2, \boldsymbol\omega_3)$. In this case, by letting $\hat{w}^\star(\boldsymbol\omega)$ and $\hat{q}(\boldsymbol\omega)$ denote the discrete-space Fourier transform (DSFT) of $w$ and $q$, respectively, the optimal solution in \eqref{L2SOL} could be defined in the DSFT domain as
\begin{equation}\label{IFILT2}
\hat{w}^\star(\boldsymbol\omega) = \left( 1 + 2 \tau |H_\vv(\boldsymbol\omega)|^2 \right)^{-1} \hat{q}(\boldsymbol\omega), 
\end{equation}
which suggests that $w^\star(\boldsymbol\omega)$ is, in fact, a linearly filtered version of $q$.

Although it is definitely possible to use \eqref{IFILT2} in practical computations, such an approach would not be recommended for two main reasons. First, FFT-based implementation of linear filtering would entail the use of periodic boundary conditions, which might not be natural for the case at hand. Second, the logarithmic complexity of FFT might still be considered to be prohibitively high (especially taking into account the fact that the filter needs to be applied $J$ times for $J$ different values of $\vv$). 

A practical alternative to an FFT-based computation of $\hat{w}^\star(\boldsymbol\omega)$ could be first to transform the  frequency response $ \left( 1 + 2 \tau |H_\vv(\boldsymbol\omega)|^2 \right)^{-1}$  back into the spatial domain, followed by truncating the impulse response thus obtained. In our experiments, the impulse responses have been truncated to a size of $7\times7\times7$ voxels, with an associated error being less than 1\% for a typical choice of the parameter $\tau$ (or, equivalently, $\mu$). The resulting finite impulse response filters have distinctive (directional) low-pass characteristics, and they can be easily applied with a linear complexity under arbitrary boundary conditions.  
  
\bibliographystyle{unsrt}
\bibliography{references}
\end{document}